%% file: main.tex
\documentclass[10pt,twocolumn,letterpaper]{article}

\usepackage[pagenumbers]{cvpr} 
\usepackage{multirow}
\usepackage{graphicx}
\usepackage{amsmath}
\usepackage{amssymb}
\usepackage{booktabs}
\usepackage{algorithm}
\usepackage{algpseudocode}
\usepackage{balance}
\usepackage{listings}
\usepackage{xfrac}
\usepackage{arydshln}

\input{math_cmd.tex}

\usepackage[pagebackref,breaklinks,colorlinks]{hyperref}

\usepackage[capitalize]{cleveref}
\crefname{section}{Sec.}{Secs.}
\Crefname{section}{Section}{Sections}
\Crefname{table}{Table}{Tables}
\crefname{table}{Tab.}{Tabs.}

\def\cvprPaperID{5966} 
\def\confName{CVPR}
\def\confYear{2023}

\begin{document}

\title{Fair Federated Medical Image Segmentation via Client Contribution Estimation}

\author{
Meirui Jiang\textsuperscript{1}, Holger R. Roth\textsuperscript{2}, Wenqi Li\textsuperscript{2}, Dong Yang\textsuperscript{2}, Can Zhao\textsuperscript{2}, \\ Vishwesh Nath\textsuperscript{2}, Daguang Xu\textsuperscript{2}, Qi Dou\textsuperscript{1,}\thanks{Corresponding authors: Qi Dou (qidou@cuhk.edu.hk) and Ziyue Xu (ziyuex@nvidia.com)}~, Ziyue Xu\textsuperscript{2,*}\\
\textsuperscript{1} The Chinese University of Hong Kong\\
\textsuperscript{2} NVIDIA \\
}
\maketitle

\begin{abstract}
How to ensure fairness is an important topic in federated learning (FL). Recent studies have investigated how to reward clients based on their contribution (collaboration fairness), and how to achieve uniformity of performance across clients (performance fairness). Despite achieving progress on either one, we argue that it is critical to consider them together, in order to engage and motivate more diverse clients joining FL to derive a high-quality global model. In this work, we propose a novel method to optimize both types of fairness simultaneously. Specifically, we propose to estimate client contribution in gradient and data space. In gradient space, we monitor the gradient direction differences of each client with respect to others. And in data space, we measure the prediction error on client data using an auxiliary model. Based on this contribution estimation, we propose a FL method, federated training via contribution estimation (FedCE), i.e., using estimation as global model aggregation weights. We have theoretically analyzed our method and empirically evaluated it on two real-world medical datasets. The effectiveness of our approach has been validated with significant performance improvements, better collaboration fairness, better performance fairness, and comprehensive analytical studies. Code is available at \url{https://nvidia.github.io/NVFlare/research/fed-ce}
\end{abstract}

\section{Introduction}
Recent development of federated learning (FL) facilitates collaboration for medical applications, given that multiple medical institutions can jointly train a consensus model without sharing raw data~\cite{rieke2020future,wu2022communication, sheller2020federated,dayan2021federated,dou2021federated,pati2022federated}. FL provides an opportunity to leverage larger and more diverse datasets to derive a robust and generalizable model~\cite{Qu_2022_CVPR,Guo_2021_CVPR}. However, it is usually difficult to pool different institutions together to train a FL model in practice. The challenges mainly lie in two aspects. 
First, it takes effort to set up and participate in federated training, medical institutions may not be sufficiently motivated to contribute to a FL study without a fair credit assignment and a fair reward allocation, i.e., \textit{collaboration fairness}~\cite{zhou2021towards}.
Second, medical data are heterogeneous in amounts and data-collection process~\cite{aubreville2020completely,fedbn,xu2021federated,Zhang_2022_CVPR}, which may lead to inferior performance for clients with either less data or a data distribution deviating from others, harming \textit{performance fairness}~\cite{qffl,mohri2019afl}.
It is critical to involve diverse datasets and improve individual prediction accuracy for building robust medical applications with low error tolerance~\cite{liu2022medical}.
Therefore, we argue that these two types of fairness need to be considered together.

Despite recent investigations on fairness-related topics, existing literature mostly addresses collaboration fairness and performance fairness separately.
For example, methods for \textit{collaboration fairness} aim to estimate client reward, by using the computation and communication cost of each client~\cite{kang2019incentive}, evaluating local validation performance~\cite{cffl}, and using cosine similarity between local and global updates~\cite{cgsv}. Meanwhile, methods for \textit{performance fairness} aim to mitigate performance disparities, by using minimax optimization to improve worst-performing clients~\cite{mohri2019afl,hu2022federated}, re-weighting clients to adjust fairness/accuracy trade-off~\cite{qffl}, or learning personalized models~\cite{ditto}.
To adequately address concerns on these two fairness, we postulate that it is desirable to consider both simultaneously, because reward estimation and model performance could essentially be coupled during training. 
Solutions on how to tackle \textit{collaboration fairness} and \textit{performance fairness} together are still under-investigated, especially in medical domain.

To tackle this problem, our insight is to estimate the contribution of each client, and further use the contribution to promote training performance. The idea is inspired by Shapley Value (SV)~\cite{shapley1997value}, a classic approach to quantify the contribution of participants in cooperative game theory. SV proposes to permute all possible subsets of participants to calculate the contribution of a certain client. Some existing works have adopted SV for estimating client reward~\cite{wei2020efficient,eci,cgsv,liu2022gtg}.
However, these methods mostly approximate SV by comparing local model updates or local model validations, which can be highly correlated with local sample numbers. A client with more samples can dominate the training, resulting in inaccurate estimation results. Therefore, finding a more accurate and robust estimation is imperative to break through this bottleneck.

In this work, we propose a novel client contribution estimation method to approximate SV by comparing a certain client with respect to all other clients. We further present a new FL algorithm, federated training via contribution estimation (\textit{FedCE}), which uses client contributions as new weighting factors for global model aggregation. Specifically, since the fundamental setting of SV is to validate if a new client contributes to all possible combinations of existing clients, to effectively and efficiently approximate it, we propose to directly measure how a certain client contributes to all remaining clients, rather than computing all possible permutations. Our contribution measurement considers both gradient and data space to quantify the contribution of each client. In gradient space, we calculate the gradient direction differences between one client and all the other clients; and in data space, we measure the prediction error on client data by using an auxiliary model, which is calculated by excluding a client's own parameters. By combining these two measurements, we are able to quantify each client's contribution for collaboration fairness, and further use this estimation to promote training for performance fairness. Our main contributions are summarized as follows:
\begin{itemize}

\item We propose a novel method for client contribution estimation to facilitate \textit{collaboration fairness}. We empirically and theoretically analyze the robustness of this estimation method under various FL data distributions.

\item We propose a novel federated learning method, \textit{FedCE}, based on the proposed client contribution estimation to help promote \textit{performance fairness}, and we theoretically analyze the model's convergence.

\item We conduct extensive experiments on two medical image segmentation tasks. The proposed FedCE method significantly outperforms several latest state-of-the-art FL methods, and comprehensive analytical studies demonstrate the effectiveness of our method.
\end{itemize}

\section{Related Works}
\subsection{Fairness in Federated Learning}
Fairness has received much attention in machine learning area, it is a broad topic that studies unintended behaviors of machine learning models~\cite{barocas2017fairness,mitchell2018prediction}. Under the setting of FL, ``individual/group fairness'', ``collaboration fairness'', and ``performance fairness'' are three most commonly studied types of fairness. The first one aims to mitigate model bias on specific protected attribute(s)~\cite{zeng2021improving,du2021fairness,galvez2021enforcing,cui2021addressing,chu2021fedfair}, the second one expects each client to receive a reward that fairly reflects its contribution~\cite{zhou2021towards,cffl}, and the third one requires uniformity of the performance distribution across clients~\cite{zhang2020fairfl,deng2020distributionally}. In this paper, we mainly focus on the latter two - ``collaboration fairness'' and ``performance fairness''.
For collaboration fairness, Kang et al.~\cite{kang2019incentive} proposed using local computation and communication cost to estimate contribution; CFFL~\cite{cffl} investigated the fairness by evaluating the validation performance on each client; and Shi et al.~\cite{shi2022fedfaim} proposed to filter out low-quality local gradients based on loss measurement.
For performance fairness, Mohri et al.~\cite{mohri2019agnostic} first proposed to optimize the performance of the single worst device by proposing a minimax optimization scheme. Later, q-FedAvg~\cite{qffl} was proposed with a more flexible optimization objective, which can be tuned based on the desired amount of fairness. Recently, Ditto~\cite{ditto} has been proposed to provide fairness by learning personalized models.
However, current studies treat these two fairness as separate problems without utilizing their underlying connection. Also, most works are validated on common benchmark datasets (e.g., MNIST and CIFAR) with arbitrary client splits. It still remains a question of how to jointly tackle collaboration and performance fairness for real-world applications in medical imaging, where client data are multi-source, highly heterogeneous, and complicated.

\subsection{Shapley Value based Client Valuation}
Shapley value (SV) is a concept measuring importance of players in cooperative game theory~\cite{shapley1997value,myerson1997game}.
Based on this, Ghorbani et al.~\cite{ghorbani2019data} proposed data SV to quantify the contribution of each data point in machine learning. Later on, Tang et al.~\cite{tang2021data} applied data SV on chest x-ray data. 
However, directly calculating SV is computationally expensive, and almost infeasible in FL with a decent number of participants. Under FL scenario, multiple studies have been performed aiming to efficiently approximate SV~\cite{jia2019towards,wei2020efficient}. For example, Kumar et al.~\cite{kumar2022towards} proposed to train linear models as proxies for client data, and used the model ensemble to approximate SV; 
Wang et al.~\cite{wang2020principled} applied SV by considering clients in an ordered sequence rather than calculating all subsets.
Song et al.~\cite{eci} proposed to approximate SV by using validation accuracy of intermediate models during federated training; CGSV~\cite{cgsv} approximated SV by using cosine similarity between local and global updates.
Liu et al.~\cite{liu2022gtg} reconstructed FL models from gradients to approximate SV instead of repeat training with different permutations.
However, these methods either require auxiliary validation data or solely rely on intermediate results. Our work aims to design a more comprehensive and practical measurement, which considers both intermediate status and actual performance without requiring extra validation data.

\section{Methods}
\subsection{Preliminary}
Let $\gD$ denote a distribution supported on a space $\gZ$, where $\gZ = \gX \times \gY$, $\gX \in \sR^d$ and $\Y$ are input and output respectively. For $N \in \sN$ clients, we have $\gD^N = \{\gD_i\}_{i=1}^N$ as the set of local client distributions, and a coalition $S \sim \gD^M$ is a subset of clients, such that $|S|=M$, where $M$ denotes number of clients in the coalition.
Let $U: \gZ \rightarrow [0,1]$ denote the utility function, where for any $S \subseteq \gZ$, $U(S)$ represents the value of this subset. For example, $U$ is typically chosen as the accuracy of an empirical risk minimizer when $S$ are training clients. We define SV as below.
\begin{definition}(Shapley Value~\cite{distSV}) 
Given a utility function $U$, a distribution $\gD$ supported on $\gZ$, and $N\in \sN$, for all client $i\in[N]$, the Shapley Value (SV) $\nu$ is defined as:
\vspace{-1mm}
$$
\nu(i ; U, \gD, N)=\underset{\substack{M \sim[N],S \sim \mathcal{D}^{M-1}}}{\mathbb{E}}[U(S \cup\{i\})-U(S)].
$$
\end{definition}
From this definition, the SV of a client is its expected marginal contribution in $U$ to a set of client coalitions $S$. To calculate SV, we need to consider all possible client coalitions, i.e., all subsets of $N$ clients. The cost will be exponentially increased with respect to the number of clients $N$, that is, $\mathcal{O}(2^N)$. Such computation is extremely expensive, even with a small number of clients. 
Therefore, it is critical to find an efficient solution for client valuation.

\begin{figure}[t]
  \centering
   \includegraphics[width=0.99\columnwidth]{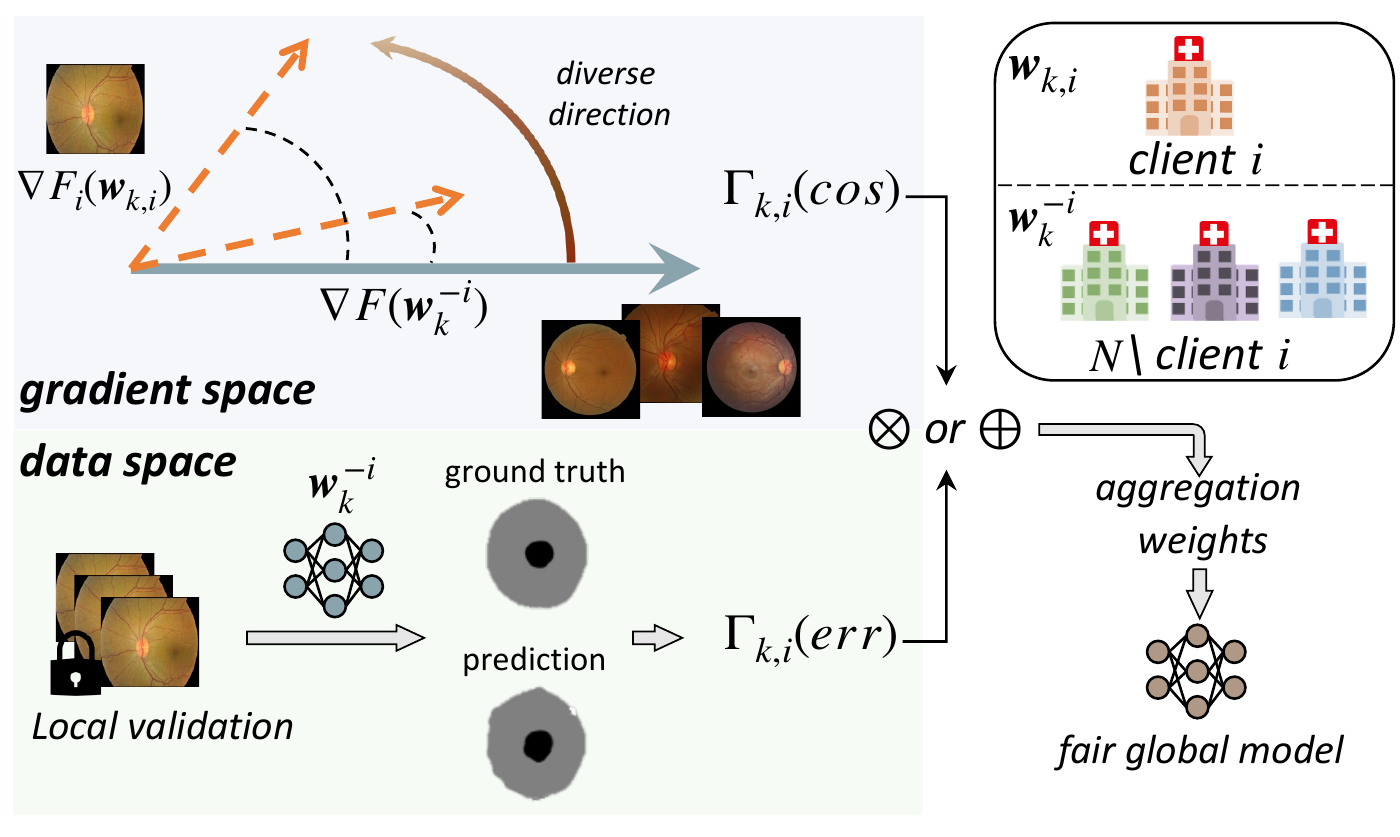}
   \vspace{-5mm}
   \caption{The proposed FedCE framework with client contribution estimation mechanism.}
   \label{fig:method}
\end{figure}

\subsection{Client Contribution Estimation}
By analyzing the SV definition, we notice that the key is to measure the value with and without a certain client with respect to all possible combinations of other clients. In other words, validate if a new client contributes (adds value) to existing clients. Therefore, we propose an efficient approximation, by directly measuring the contribution of client $i$ to others ($N\setminus\!\{i\}$). We define our new value $\hat{\nu}$ as:
\begin{equation}
\label{eq:sv_our}
\begin{aligned}
\hat{\nu}(i ; \Gamma, \gD, N)&=\underset{S\sim \mathcal{D}^{N}}{\mathbb{E}}\left[\Gamma_i(S\setminus\{i\},\{i\})\right] 
\\
&=\Gamma_{i}\left(N\setminus\{i\},\{i\}\right),
\end{aligned}
\end{equation}
where ${\Gamma_i}$ is our proposed function to measure the contribution of client $i$. Different from common implementations of the utility function $U$ using accuracy, we propose a more comprehensive way to measure the contribution by considering both gradient and data space, as shown in Fig.~\ref{fig:method}. 

We first introduce the measurement in gradient space by using cosine similarity (cos). For client $i$ at $k$-th round:
\begin{equation}
\label{eq:cos}
\Gamma_{k,i}(cos) \triangleq 1 - cos\left(\nabla F_i(\bs w_{k,i}), \nabla F(\bs w_k^{-i})\right),
\end{equation}
where $\nabla F_i(\bs w_{k,i})$ denotes the local client gradient, which is calculated by differences between global model parameter $\bs w_k$ and local model parameter $\bs w_{k,i}$. $\nabla F(\bs w_k^{-i}) = \sfrac{(\nabla F(\bs w_k) -  p_i \nabla{F_i(\bs w_{k,i})})}{(1-p_i)}$ is aggregated gradients excluding client $i$ and $p_i \geq 0$ denotes client importance (e.g., proportional to client sample number). Global gradients is denoted by $\nabla F(\bs w_k)= \sum_{i=1}^N {p_i} \nabla F_i(\bs w_{k,i})$ where $\sum_{i=1}^N p_i=1$. Then we further normalize the term, i.e., $\Gamma_{k,i}(cos) = \sfrac{\Gamma_{k,i}(cos)}{\sum_{i=1}^N \Gamma_{k,i}(cos)}$ (for ease of notation, we reuse $\Gamma_{k,i}(cos)$), to ensure the summation over clients adds up to 1.
$\Gamma_{k,i}(cos)$ quantifies the contribution by measuring the optimization direction of client $i$ compared with others. In particular, if the cosine similarity between client $i$ and others is close to 1, $\Gamma_{k,i}(cos)$ becomes 0, indicating this client does not represent a new direction information. Hence, removing client $i$ will have little impact on the global model's update direction. We argue that it is important to capture the large data variety and heterogeneity in FL for training a robust and generalizable global model. Therefore, we assign more weight to clients presenting different gradient directions.

However, as an indication in gradient space, a gradient direction that is different from others may not be sufficient to fully measure the contribution from a certain client to the overall FL model performance. Consequently, we further propose a measurement in data space by calculating the model error on the clients' data. Similar to the client exclusion setting in gradient space, we calculate the aggregated model parameters by excluding client $i$ at $k$-th round, i.e., $\bs w_{k}^{-i} = \sfrac{(\bs w_k - p_i \bs w_{k,i})}{(1-p_i)}$. Then we validate this new model on client $i$'s data samples:
\begin{equation}
\label{eq:err}
\Gamma_{k,i}(err) \triangleq  \gE(\hat{\gD}_i; \bs w_k^{-i}),
\end{equation}
where $\gE(\hat{\gD}_i; \bs w_k^{-i})$ denotes the error on the empirical distribution $\hat{\gD}_i$, here we use the validation samples in client $i$, i.e., the error is the full performance score
``1'' minus the obtained validation performance. We also normalize this term to ensure the summation over clients is 1. The intuition of measuring error is that, if a client presents a new data distribution, using model parameters from other clients only may not result in a good performance. Otherwise, if $\bs w_{k}^{-i}$ already achieves low error on $\hat{\gD}_i$, incorporating updates from client $i$ may not improve the overall performance significantly. By further adding this assessment, we complement the previous findings in gradient space and better determine the contribution from a client to the overall model performance.

To combine these two factors for the contribution estimation, 
we choose multiplication and summation as two alternative mechanisms. We name them as $\Gamma_{k,i}^m$ and $\Gamma_{k,i}^s$, respectively, and formulate them as:

\begin{equation}
\label{eq:comb_ways}
\left\{\begin{array}{ll}
\Gamma_{k,i}^m &= \Gamma_{k,i} (cos) \times \Gamma_{k,i}(err) \\
\Gamma_{k,i}^s &= \Gamma_{k,i} (cos) + \Gamma_{k,i}(err).
\end{array}\right.
\end{equation}
These contribution estimation terms are calculated at each communication round, and by accumulating them over all $K$ rounds, we can derive the final contribution estimations. We evaluate both combinations in our experiments.

\subsection{Federated Training via Contribution Estimation -- FedCE}
From the proposed formulation of client contribution, it is natural to consider using these contribution estimation results to further improve federated training. In this regard, we propose a new federated algorithm, FedCE, by using client contributions as weighting factors for global model aggregation. 
Instead of using the standard federated averaging (FedAvg) weight $p_i$, which is typically proportional to the data amount of each client~\cite{fedavg}, we use our estimated client contribution as the new weight. 

There are two benefits of using client contributions to promote fairness. First, client contributions are more comprehensive and fair than the standard weights based on sample numbers.  The standard weights is a weak representation of client data distribution and could be vulnerable to data manipulations, such as increasing number by repeating.
Second, contribution-based aggregation encourages the global model to cover a wide range of data distributions. Distributions sharing a common pattern are easy to fit. In contrast, clients with rare distribution are usually under-represented in training, which is a potential driver of model performance unfairness. Our contribution mechanism helps promote training on these clients, because they present different data information. As a result, this will facilitate the performance fairness of the global model.

Taking the multiplication-based combination as an example, at the $k$-th round, we calculate the aggregation weights as follows:
\vspace{-1mm}
\begin{equation}
\label{eq:rho_m}
\bs\rho_k^m = \frac{1}{Z_k}\left[ \sum\nolimits_k\Gamma_{k,0}^{m},\cdots,\sum\nolimits_k\Gamma_{k,N}^{m}\right],
\vspace{-1mm}
\end{equation}
where $Z_k=k\sum_{i=1}^N\sum_k\Gamma_{k,i}^m$ is a normalization factor to ensure $\sum_{i=1}^N\rho_{k,i}^m=1$. Then we obtain the global model by using the new weights to aggregate local client gradients:
\vspace{-1mm}
\begin{equation}
\label{eq:aggregation}
\bs w_{k+1} \leftarrow \bs w_k - \eta\sum\nolimits_{i=1}^N \rho_{k,i}^m \cdot \nabla F_i(\bs w_{k,i}).
\vspace{-1mm}
\end{equation}
\noindent Similarly, for the summation-based combination (defined in Eq.~(\ref{eq:comb_ways})), we compute the aggregation weights of $\bs \rho_k^s$ by replacing $\Gamma_{k,i}^m$ with $\Gamma_{k,i}^s$. 
We present the full algorithm in Algorithm~\ref{alg:algorithm}. The final outputs are global model and client contribution estimations.
Our aggregation weight is dynamical, and it considers all the historical information.
Note that our method does not require any extra training compared with FedAvg~\cite{fedavg}. Furthermore, the contribution estimation is performed locally, which helps reduce the communication burden and potential risk of information leakage.

\begin{algorithm}[tb]
\caption{Our proposed method FedCE}
\label{alg:algorithm}
\textbf{Input}: communication rounds $K$, number of clients $N$, local datasets $\{\hat{\gD}_i\}_{i=1}^N$, learning rate $\eta$, local steps $\{\kappa_i\}_{i=1}^N$.\\
\textbf{Output}: final global model $\bs w_K$, contributions $\{\rho_{K,i}\}_{i=1}^N$.
\begin{algorithmic}[1] 
\State Initialize server model $\bs w_0$ 
\For{$k=1,\cdots, K-1$}
\State \textit{Server}: $\bs w_{k,i}^0 \leftarrow \bs w_k$ \Comment{distribute global model $\bs w_k$}
\For{\textit{Client} $i=1,2,\cdots, N$ {in parallel}} 
\State $\nabla F(\bs w_k) = \bs w_k - \bs w_{k-1}$
\For{$j=1,2,\cdots, \kappa_i$} \Comment{client training}
\State $\bs w_{(k,i)}^{j+1}\leftarrow \bs w_{k,i}^j -\eta \nabla F_i(\bs w_{(k,i)}^j)$
\EndFor
\State $\nabla F_i (\bs w_{k,i}) = \bs w_{k,i}^{\kappa_i} - \bs w_{k,i}^0$
\State  $\nabla F(\bs w_k^{-i}) = \frac{\nabla F(\bs w_k) -  \rho_{(k\!-\!1,i)} \nabla{F_i(\bs w_{k,i})}}{1-\rho_{(k\!-\!1,i)}}$ 
\State $\Gamma_{\!k,i}(cos) \!\!=\!\! 1 \!\!-\! cos(\nabla \!F_{\!i}(\!\bs w_{k,i}),\!\! \nabla \!F(\!\bs w_k^{\!-i}))\!\!$
\Comment{Eq.(\ref{eq:cos})}
\State $\bs w_{k}^{-i} = \frac{(\bs w_k - \rho_{(k\!-\!1,i)} \bs w_{k,i})}{(1-\rho_{(k\!-\!1,i)})}$.
\State $\Gamma_{k,i}(err) =  \gE(\hat{\gD}_i; \bs w_k^{-i})$ \Comment{Eq.(\ref{eq:err})}
\State {calculate} $\rho_{k,i}$ \Comment{Eq.(\ref{eq:rho_m})}

\State \textbf{return} $\bs w_{k,i}^{\kappa_i}, \rho_{k,i}$ \Comment{send client model and contribution to server}
\EndFor
\State \textbf{Server}: $\bs w_{k+1} \leftarrow \sum_{i=1}^N \rho_{k,i} \bs w_{k,i}^{\kappa_i}$ 
\EndFor 
\State \textbf{return} $\bs w_K$, $\{\rho_{K,i}\}_{i=1}^N$
\end{algorithmic}
\end{algorithm}

\subsection{Theoretical Analysis for FedCE}
Since our contribution value is naturally parameterized by the underlying data distribution $\gD$, it is helpful to investigate how the value will change if the data distribution changes.
Here we formally quantify the value differences under distributional shift by presenting an upper bound.
\begin{theorem}
\label{thm:sv_dist}
Let $\Gamma$ be $\gB$-Lipschitz stable with respect to $\gZ$. Suppose $\gD_s$ and $\gD_t$ are two distributions over $\gZ$. Then, for all $N \in \sN$ and all $i \in \gZ$,
$$
\left|\hat{\nu}\left(i ;\Gamma, \mathcal{D}_{s},N \right)-\hat{\nu}\left(i; \Gamma, \mathcal{D}_{t}, N\right)\right| \leq 2N \gB \cdot W_{1}\left(\mathcal{D}_{s}, \mathcal{D}_{t}\right),
$$
\end{theorem}
where $W_1$ denotes the Wasserstein distance between two distributions. This theorem measures the values changes under two different data distributions. If two distributions are similar, then similar values should be obtained. While for two different distributions, we can bound the difference in terms of the Wasserstein distance. For more details and proofs, please refer to Appendix.~\ref{app:sec:proof_dis}.

Then we analyze the convergence behavior of our method. To complete the analysis, we adopt assumptions on local function smoothness and gradient variance, which are classically used in optimization literature~\cite{li2019convergence,fedprox,fedadam,scaffold,tong2020effective}. We present our results below.
\begin{theorem}
\label{thm:convergence}
Assume the objective function is Lipschitz smooth and gradient variance is bounded. In the $k$-th round for $k \in [0, K-1]$, let $\beta_{(k,i)}$ and $\delta_{(k,i)}$ be factors relate to variance bounding for client $i$, $L$ be factor of smoothness, and $\eta$ be learning rate, when $\eta L - 1 \geq 0$, we have:
\begin{align*}
&F\left(\bs {w}_{k+1}\right)-F\left(\bs {w}_{k}\right)
\\ &\leq
\left(\frac{\eta}{4}(2\eta L + \sum_{i=1}^{N} p_{i} \rho_{(k,i)} \eta A_{(k,i)}) - \eta\right)\|\nabla F(\bs w_k)\|^2.
\end{align*}
\end{theorem}
where $A_{(k,i)}\!\triangleq\!\eta{\sqrt{\kappa_{(k,i)}
}(\kappa_{(k,i)}\!-\!1)}\beta_{(k,i)}^2 \delta_{(k,i)}$ is a variable that relates to $\beta_{(k,i)}^2$, $\delta_{(k,i)}$ and local iteration steps $\kappa_{(k,i)}$.
The theorem analyzes the upper bound of the convergence with our method in the context of global model update. Furthermore, we present another corollary to determine the upper bound on our aggregation factor $\rho_{(k,i)}$.
\begin{corollary}
For $k \in [0, K-1]$, considering the $A_{(k,i)}$-dominated convergence, the model converges when 
$$
\rho_{(k,i)} \leq \frac{4A^{1/2}_{(k,i)}}{(A_{(k,i)}-2L)} = \mathcal{O}(\frac{1}{\sqrt{A_{(k,i)}}}).
$$
\end{corollary}
The proof can be found in Appendix~\ref{app:sec:proof_conv}.
According to this corollary, we could promote the convergence by minimizing the upper bound, that is, increasing $A_{(k,i)}$. 
Specifically, this term contains four terms. For $\eta$ and $\kappa$, it is intuitive that increasing the learning rate or local iteration steps can increase model convergence speed at an early stage. However, it may let the model trap into a local optimum or suffer large client drifts when data are heterogeneous~\cite{scaffold}. For the term $\beta$, it can be converted to a form related to $\delta$. So we discuss $\delta$ here. The term $\delta$ is a variable in one of our assumptions, which can be written as $\sfrac{\left\|\sum_{s=0}^{\lambda-1} \nabla F_{i}\left(\bs {w}_{(k, i)}^{s}\right)\right\|^{2}}{(\sum_{s=0}^{\lambda-1}\left\|\nabla F\left(\bs {w}_{k}\right)\right\|^{2})}$. This term quantifies the percentage of local gradients over
the global(aggregated) gradients. That is, to increase $\delta_{(k,i)}$, we need to weigh more on local gradients
from client $i$. Since the client with boundary data or different distribution is under-represented during training, which harms the overall convergence. We need to assign higher weights to promote training on this kind of client, thus improving convergence. This well matches our contribution estimation method, i.e., allocating higher weight to clients presenting different information in gradient space or suffering high error on local data when their gradient is excluded.

\section{Experiment}
Our method is evaluated on two medical image segmentation tasks: retinal fundus image segmentation~\cite{orlando2020fundus} and prostate MRI segmentation~\cite{prostate2014evaluation}. We compare our method with other methods on segmentation performance, performance fairness, and collaboration fairness. We also conduct in-depth analyses on our method, including convergence speed, robustness to free riders and distribution changes, and effectiveness of each component. 
For more results, please refer to Appendix~\ref{app:sec:more_exp}.

\subsection{Experimental Settings}
\begin{figure}[t]
\centering
\includegraphics[width=0.99\columnwidth]{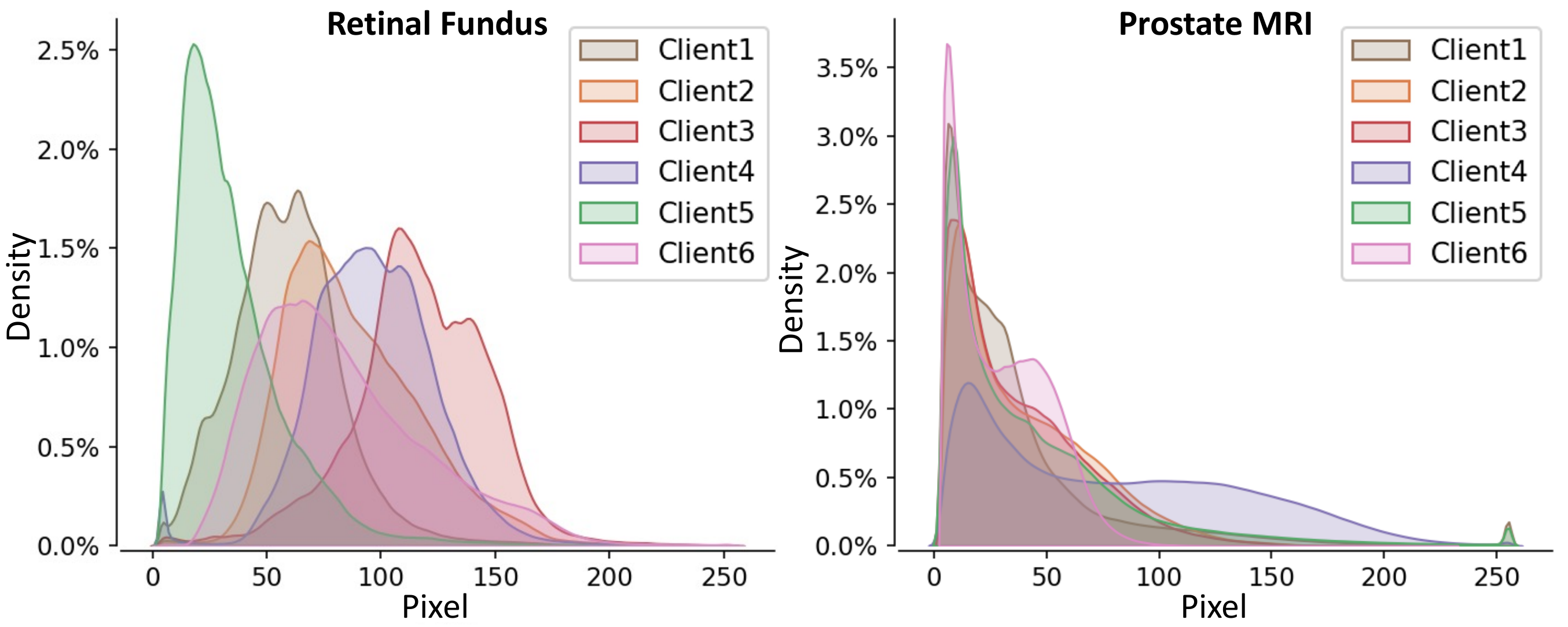} 
\vspace{-3mm}
\caption{Pixel intensity distributions. Left denotes samples from retinal fundus dataset and right ones are from prostate MRI.}
\label{fig:data_dist}
\end{figure}

\noindent\textbf{Datasets.}
We evaluate our approach on two medical image segmentation datasets: the prostate MRI dataset from 6 institutions~\cite{liu2020ms,prostate2014evaluation,prostate2015computer,isbi}, and the retinal fundus dataset from 6 different sources~\cite{fumero2011fundus,sivaswamy2015fundus,almazroa2018retinal,orlando2020fundus}. Each institution/source is treated as a single client, and the data is randomly split into training, validation, and testing sets with a ratio of 0.5, 0.25, and 0.25 for each client. All images are resized to 256$\times$256. 
Note that the data collected from different medical centers present realistic heterogeneous distributions, due to varying local devices and imaging protocols. As shown in Fig.~\ref{fig:data_dist}, the retinal fundus dataset has lower data distribution similarity among clients, while the prostate dataset has a relatively higher data similarity.
\\\textbf{Evaluation metrics.}
To comprehensively evaluate our approach, we adopt four different metrics. We use the Dice coefficient (Dice) to evaluate segmentation results. We use the Pearson correlation and Euclidean distance to measure the performance fairness, and further add cosine similarity to evaluate the accuracy of contribution estimation. Following the fairness definition from~\cite{qffl,ditto}, we also calculate the standard deviation of test performance among clients.
\\\textbf{Implementation details.}
In our implementation, all methods use the same training settings. The loss function is dice loss~\cite{milletari2016v}, and the optimizer is Adam with $\beta=(0.9,0.99)$. The learning rate is $1e-3$ and the batch size is set to 8. We trained for 200 federated rounds to ensure that the model converged steadily, and the local update epoch is set to 1.
\begin{table*}[t!]
    \renewcommand\arraystretch{1.2}
    \centering
        \caption{\small{Performance comparison using Dice score on image segmentation datasets of retinal fundus images and prostate MRI.}}
        \centering
        \vspace{-3mm}
        {
        \scalebox{0.88}{
        \setlength\tabcolsep{4pt}
        \begin{tabular}{c|cccccccc|cccccccc}
    \toprule
             {Task}&\multicolumn{8}{c|}{Retinal Fundus Segmentation}&\multicolumn{8}{c}{Prostate MRI Segmentation}
              \\\cline{0-16}
            Client & 1 & 2 & 3 & 4 & 5 & 6 & Avg. & Std.
            &1 & 2 & 3 & 4 & 5 & 6 &Avg. & Std. \\
            \hline
            Standalone &
            86.69 & 85.51 & 86.21 & 89.91 & 
            79.77 & 90.98& 
            86.51 & 3.95 & 
            91.23 & 84.59 & 87.57 & 87.37 & 
            86.70 & 89.25 & 
            87.79 & 2.26\\
            
            \hline
            FedAvg &
            81.34 & 85.21 & 83.28 & 88.16 & 
            40.81 & 90.79 & 
            78.27 & 18.66 & 
            
            91.10 & 84.59 & 89.02 & 89.09 & 
            83.87 & \textbf{89.27} & 
            87.82 & 2.90\\
            
            q-FedAvg &
            86.24 & 86.97 & 87.37& 89.13 & 
            44.68 & 90.72 & 
            80.85 & 17.80 & 
            
            90.94 & 85.60 & 89.28 & 89.18 & 
            84.27 & 88.67 & 
            87.99 & 2.52\\
            
            CFFL &
            85.72 & 86.29 & 86.96 & 88.62 & 
            41.12 & 90.16 & 
            79.81 & 19.02 & 
            
            91.01 & 85.49 & 89.24 & 88.98 & 
            82.11 & 88.17 & 
            87.50 & 3.20\\
            
            FedCI &
            87.02 & 86.93 & 87.35 & 88.53 & 
            40.99 & 90.22 & 
            80.17 & 19.24 & 
            
            91.21 & 85.40 & 89.49 & 88.37 & 
            83.96 & 88.49 & 
            87.82 & 2.68\\
            
            CGSV &
            83.46 & 85.57 & 85.47 & 88.48 & 
            33.79 & \textbf{91.01} & 
            77.96 & 21.80 & 
            
            91.15 & 84.90 & 89.27 & 88.09 & 
            83.47 & 89.16 & 
            87.67 & 2.91\\
            
            \hline

            FedCE (Multi.) &
            86.73 & \textbf{87.45} & 87.51 & 89.26 & 
            \textbf{57.30} & 90.25 & 
            \textbf{83.08} & \textbf{12.70} & 
            
            \textbf{91.43} & \textbf{85.79} & 89.21 & 89.13 & 
            \textbf{85.68} & 88.62 & 
            \textbf{88.31} & \textbf{2.22}\\         
            
            FedCE (Sum.) & 
            \textbf{87.22} & 87.36 & \textbf{87.93} & \textbf{89.66} & 
            54.42 & 90.92 & 
            82.92 & 14.03 & 
            
            91.18 & 85.54 & \textbf{89.59} & \textbf{89.22} & 
            84.99 & 88.79 & 
            88.22 & 2.43 \\
            
       \bottomrule
        \end{tabular}
    }}
    \vspace{-1mm}
    \label{table:dice_res}
\end{table*}
\\\textbf{Baseline methods.}
We compare our approach with state-of-the-art (SOTA) methods targeting collaboration fairness and performance fairness, including: q-FedAvg~\cite{qffl}, a method to learn fair performance distribution; CFFL~\cite{cffl}, a method for collaboration fairness by evaluating local participant's validation accuracy; FedCI, a method we extend from a client valuation method CI~\cite{eci}, by using the valuation results as aggregation weights; CGSV~\cite{cgsv}, a method quantifying client reputation based on SV. Furthermore, we also compared with the FedAvg~\cite{fedavg} and Standalone (local training and evaluation on each client's own data).

\subsection{Experimental Results}   
\noindent\textbf{Segmentation performance.} We first present the comparison on segmentation performance. Table~\ref{table:dice_res} lists all the quantitative results for two segmentation tasks, including performance on each client, the average performance, and the standard deviation across clients.The goal of performance fairness methods is to lower variance while maintaining the average performance.
From the table, it can be observed that client 5 suffers a significant performance drop across FL methods due to less similar data distributions in retinal dataset. 
Since the compared methods may not specifically consider such large heterogeneity in training data, their performance on client 5 is lower than FedAvg and suffer a higher standard deviation. For other clients, these methods present higher or comparable results. Compared with them, our approaches achieve higher overall performance with an increase of 2.23\% and 2.07\% and make improvements on most clients (5 over 6). And the performance variance of our method is significantly lower than others (with a decrease of 5.10). 
For prostate MRI, most methods achieve higher average performance and lower variance than FedAvg. And our methods outperform all SOTA methods in terms of overall performance and variance.
\begin{table}[t]
    \renewcommand\arraystretch{1.2}
    \centering
        \caption{\small{Fairness comparison with our method and others. We use Pearson correlation ($\uparrow$) and Euclidean distance ($\downarrow$) as metrics. Value in parentheses denotes the p-value.}}
        \centering
        \vspace{-3mm}
        \scalebox{0.85}{
        \setlength\tabcolsep{2pt}
        \begin{tabular}{c|c|c|c|c}
        \toprule
         Task & 
         \multicolumn{2}{c|}{\begin{tabular}[c]{@{}c@{}}Retinal Fundus \\Segmentation\end{tabular}}&
         \multicolumn{2}{c}{\begin{tabular}[c]{@{}c@{}}Prostate MRI \\Segmentation\end{tabular} }
         \\\cline{0-4} 
        \hline
        Metric & \begin{tabular}[c]{@{}c@{}}Pearson\\Correlation\end{tabular} & \begin{tabular}[c]{@{}c@{}}Euclidean\\Distance\end{tabular} & \begin{tabular}[c]{@{}c@{}}Pearson\\Correlation\end{tabular} & \begin{tabular}[c]{@{}c@{}}Euclidean\\Distance\end{tabular} \\ \hline
        FedAvg~\cite{fedavg} & 88.82 \small{(1.8$e^{\!-2}$)} & 38.94 & 88.67 \small{(1.9$e^{\!-2}$)} & 3.31 \\
        q-FedAvg~\cite{qffl} & 87.02 \small{(2.4$e^{\!-2}$)} & 38.96 & 91.69 \small{(1.0$e^{\!-2}$)} & 2.61 \\
        CFFL~\cite{cffl} & 84.53 \small{(3.4$e^{\!-1}$)} & 38.96 & 92.47 \small{(8.3$e^{\!-3}$)} & 2.46\\
        FedCI~\cite{eci} & 85.69 \small{(2.9$e^{\!-2}$)}& 48.42 & 93.57 \small{(6.1$e^{\!-3}$)} & 2.25 \\ 
        CGSV~\cite{cgsv} & 87.47 \small{(2.3$e^{\!-2}$)} & 49.47 & 87.71 \small{(2.2$e^{\!-2}$)} & 3.45\\ \hline
        
        FedCE (Multi.) & 88.94 \small{(1.8$e^{\!-2}$)} & \textbf{24.57} & \textbf{98.25} \small{(4.6$e^{\!-4}$)} & \textbf{1.09}\\
        
        FedCE (Sum.)& \textbf{89.11} \small{(1.7$e^{\!-2}$)} & 24.69 & 97.15 \small{(1.2$e^{\!-3}$)} & 1.41\\ \hline
        
       \bottomrule
        \end{tabular}
    }
    \vspace{-1mm}
    \label{table:fairness_res}
\end{table}
\\\textbf{Performance fairness.}
One aim of our study is to improve performance fairness, which we evaluate and compare our method with others on fairness metrics. Besides comparing the standard deviation of performance in Table~\ref{table:dice_res}, we further consider using a scaled Pearson correlation and Euclidean distance, which are also adopted in~\cite{cgsv,jia2019towards}.  The results of our fairness comparison are presented in Table~\ref{table:fairness_res}, where we calculate the Pearson correlation and Euclidean distance between the test Dice scores of standalone and other methods.
Our methods consistently achieve a high degree of fairness compared to others, as indicated by the higher correlation value and lower distance to the Standalone results.
The p-value of our results is smaller than 0.05 and also lower than others. 
Notably, for prostate segmentation, we observe that methods with better fairness than FedAvg also achieve higher accuracy and smaller variance, highlighting the importance of fairness metrics in performance evaluation. 
\begin{table*}[htbp]
    \renewcommand\arraystretch{1.2}
    \centering
        \caption{\small{Client contribution estimation comparison by comparing the results of leave-one-out with that of our method and others. We use Pearson correlation ($\uparrow$), Euclidean distance ($\downarrow$), and cosine similarity ($\uparrow$). Value in parentheses denotes the p-value.}}
        \centering
        \vspace{-3mm}
        \scalebox{0.88}{
        \setlength\tabcolsep{4pt}
        \begin{tabular}{c|c|c|c|c|c|c}
        \toprule
         Task & 
         \multicolumn{3}{c|}{\begin{tabular}[c]{@{}c@{}}Retinal Fundus Segmentation\end{tabular}}&
         \multicolumn{3}{c}{\begin{tabular}[c]{@{}c@{}}Prostate MRI Segmentation\end{tabular} }
         \\\cline{0-4} 
        \hline
        Metric & \begin{tabular}[c]{@{}c@{}}Pearson\\Correlation\end{tabular} & \begin{tabular}[c]{@{}c@{}}Euclidean\\Distance\end{tabular} & \begin{tabular}[c]{@{}c@{}}Cosine\\Similarity\end{tabular} & \begin{tabular}[c]{@{}c@{}}Pearson\\Correlation\end{tabular} & \begin{tabular}[c]{@{}c@{}}Euclidean\\Distance\end{tabular}  & \begin{tabular}[c]{@{}c@{}}Cosine\\Similarity\end{tabular} \\ \hline
        FedAvg~\cite{fedavg} & -39.76 \small{(4.4$e^{-1}$)} & 0.55 & 0.26 & 
        3.01 \small{(9.5$e^{-1}$)} & 0.62 & 0.52\\
        q-FedAvg~\cite{qffl} & 63.28 \small{(1.8$e^{-1}$)} & 0.31 & 0.57 & 
        63.35 \small{(1.8$e^{-1}$)} & 0.59 & 0.57\\
        CFFL~\cite{cffl} & 0.90 \small{(9.9$e^{-1}$)}& 0.45 & 0.47 & 
        75.44 \small{(8.3$e^{-2}$)} & 0.49 & 0.74\\
        FedCI~\cite{eci} & -12.36 \small{(8.2$e^{-1}$)}& 0.37&  0.53 &
        -0.31 \small{(1.0$e^{0}$)} & 0.61 & 0.53 \\ 
        CGSV~\cite{cgsv} &  -44.50 \small{(3.8$e^{-1}$)}& 0.57 & 0.22 & 
        -1.85 \small{(9.7$e^{-1}$)} & 0.63 & 0.50 \\ \hline
        
        FedCE (Multi.) & 94.93 \small{(3.8$e^{-3}$)}& \textbf{0.17} & \textbf{0.82} & 
        {93.12} \small{(6.9$e^{-3}$)} & \textbf{0.49} & \textbf{0.75}\\
        
        FedCE (Sum.)& \textbf{96.34} \small{(2.0$e^{-3}$)}& 0.22 &  0.73 & 
        \textbf{93.53} \small{(6.1$e^{-3}$)} & 0.53 & 0.69\\ \hline
        
       \bottomrule
        \end{tabular}
    }
    \label{table:loo_measure}
\end{table*}
\begin{figure*}[t]
\centering
\includegraphics[width=0.95\textwidth]{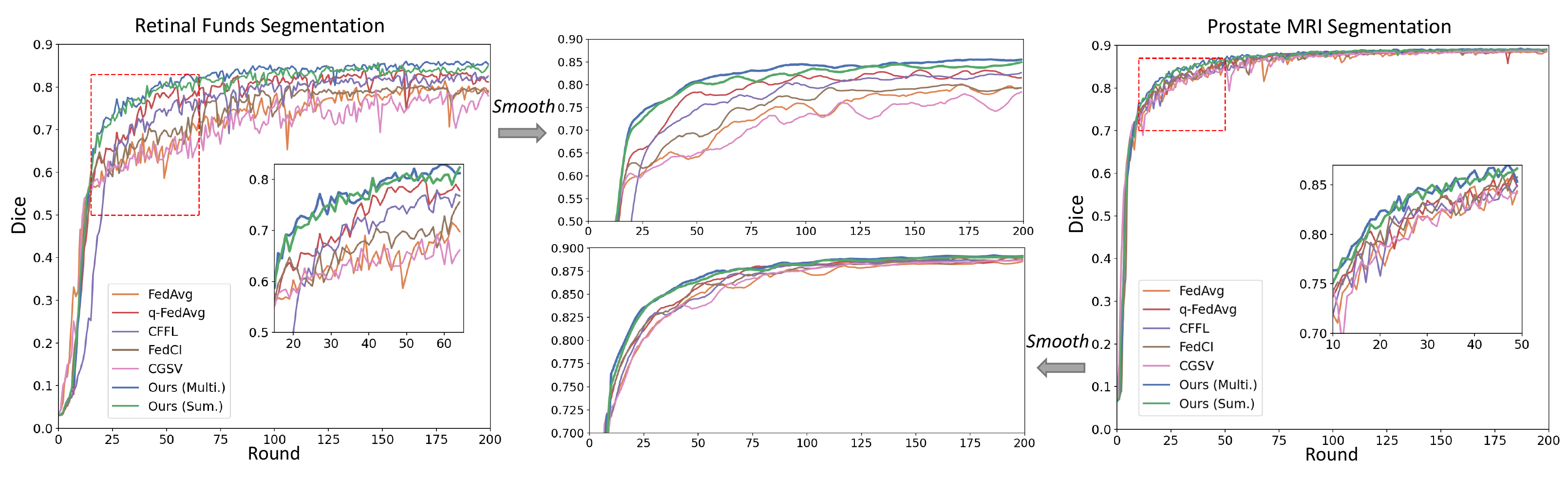} 
\vspace{-6mm}
\caption{Convergence analysis in terms of validation Dice score with the different number of communication rounds.}
\label{fig:convergence}
\end{figure*}
\\\textbf{Collaboration fairness.}
For collaboration fairness, our proposed method provides an indication of reward/profit distribution by measuring the contribution of clients. For methods in comparison, except FedAvg, the aggregation weights of others are also dynamic during training. So we take the aggregation weights as client contributions and perform the comparison. 
To obtain the "ground-truth" of a client's contribution, we conduct leave-one-out experiments, a popular and reliable way for data valuation~\cite{ghorbani2019data}. This approach assesses how much performance we will lose if we remove a certain client. After obtaining leave-one-out results, we compare them with contributions estimated by our approach and other methods.
Table~\ref{table:loo_measure} presents all the comparison results using three different metrics. It can be observed that, our methods achieve a higher correlation and cosine similarity, as well as lower distance compared with others. We notice that FedAvg presents a low correlation value, it is reasonable because the proportion of sample number may not correlate well with performance improvements.
These three metrics comprehensively validate the accuracy of our client contribution estimation.

\subsection{Analytical Studies}   
\begin{figure}[t]
\centering
\includegraphics[width=0.99\columnwidth]{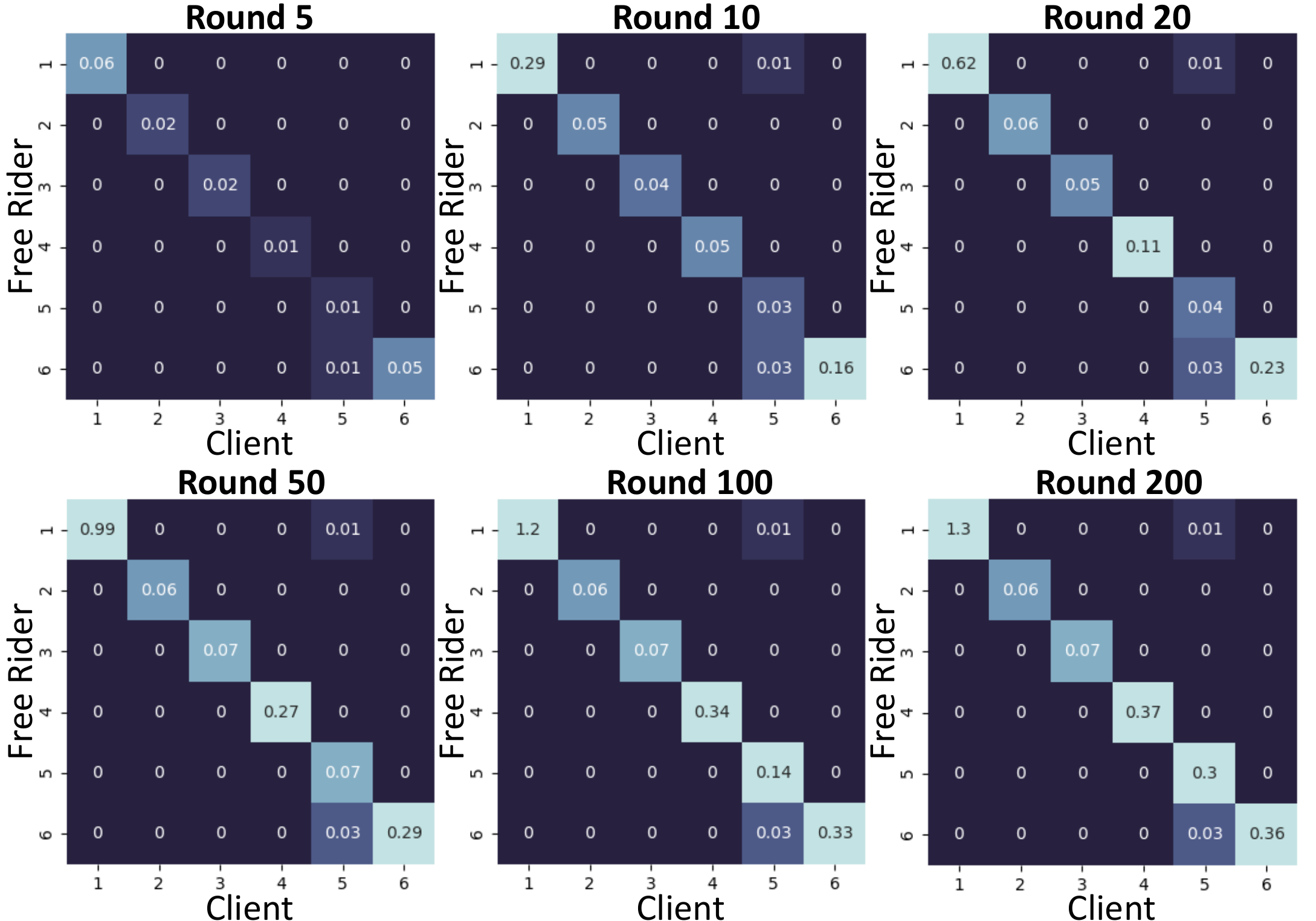} 
\vspace{-3mm}
\caption{Free rider study. Each row denotes an independent federated training procedure, and the y-axis indicates which client is the free rider. A free rider is detected with a high value.}
\label{fig:freeriders}
\end{figure}

We further conduct in-depth analytical studies to investigate key properties of our method, including: i) the convergence speed; ii) the robustness against free riders; iii) the robustness against distribution changes; and iv) the contribution of each measurement metric.
\\\textbf{Convergence speed.}
We first show the average validation Dice score across clients per communication round for different FL methods. 
As shown in Fig.~\ref{fig:convergence}, it can be observed that the curves of our methods converge to higher performance with faster speed than compared methods. This attributes to our contribution-based aggregation, which involves diverse gradients to promote global model optimization on the overall data distribution. This observation also validates our theoretical analysis that using contribution estimation to aggregate models in FL helps promote convergence. In addition, we applied Savitzky–Golay filter~\cite{savitzky1964smoothing} to smooth the curves to better present the overall trend.
\\\textbf{Robustness against free riders.}
We then study a situation where a ``free rider'' joins the FL: if a client does not have enough data to participate in FL, it may cheat by repeating one image several times, and try to obtain the global model for its own use. However, in such case, the free rider has almost no contribution to the global training, and it should not enjoy such ``free lunch''~\cite{lyu2020threats}. We hereby consider identifying the free rider by calculating a new value, which is the multiplication of local-global gradients cosine similarity and local-global model error difference. Note that these gradients and models are naturally generated during our method training. The results are shown in Fig.~\ref{fig:freeriders}, we present snapshots on six communication rounds. Each row denotes an independent federated training, and the y-axis shows which client is the free rider. It can be observed that free riders are detected at very early stages, i.e., within 10 rounds, and as training goes on, the results become more significant. Note that when client 6 is the free rider, even though client 5 has a relatively high value of 0.03 at round 10, client 6 has a significantly higher value (over $5$ times) than client 5. Our results present the potential of identifying free riders at an early stage to save time and development costs in real-world practice.
\begin{figure}[t]
\centering
\includegraphics[width=0.95\columnwidth]{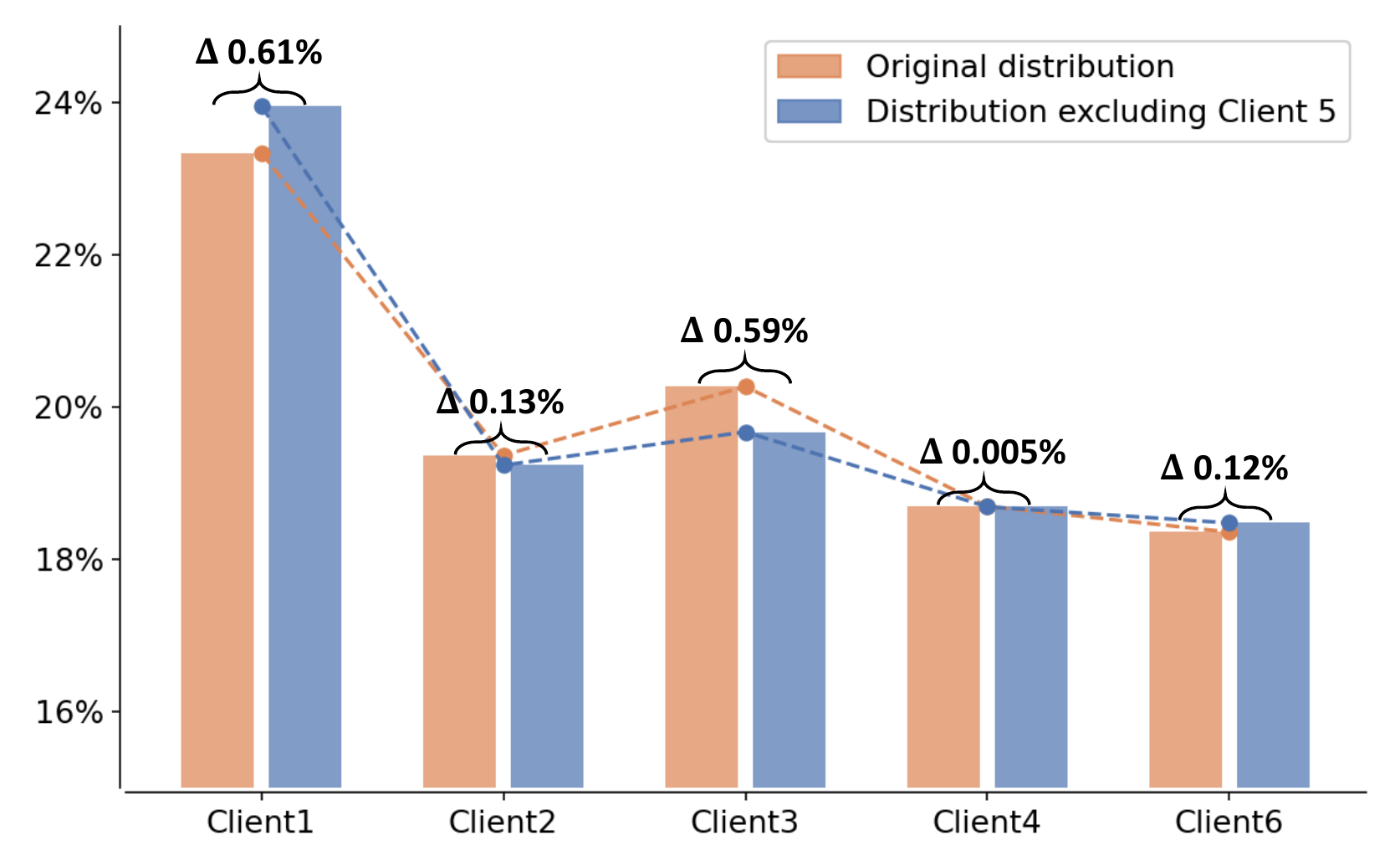} 
\vspace{-4mm}
\caption{Study to validate the distribution shift robustness of our contribution estimation metric. The y-axis denotes the relative weight percentage, and $\Delta$ denotes the differences.}
\label{fig:dist_robust}
\vspace{-2mm}
\end{figure}
\\\textbf{Robustness against distribution changes.}
For client contribution estimation, we may expect that the relative value among clients should be robust to distribution changes. Therefore, we form two distributions to investigate the estimation robustness of our method. We notice the overall clients' distributions differ a lot by including/excluding client 5 on retinal fundus dataset. We use ``original distribution'' to denote all 6 clients and ``distribution excluding Client 5'' to denote the collection of other 5 clients. We present the estimation value in Fig.~\ref{fig:dist_robust}. The five estimation results from ``original distribution'' are re-normalized. Our metric presents similar estimation values for these 5 clients, even though the overall distributions are different. The value changes are smaller than 1\%. And two different estimations have a similar trend, as shown by the curves. This study also empirically validates our theorem~\ref{thm:sv_dist} for the upper bound of value changes under distribution shift.
\begin{figure}[t]
\centering
\includegraphics[width=0.95\columnwidth]{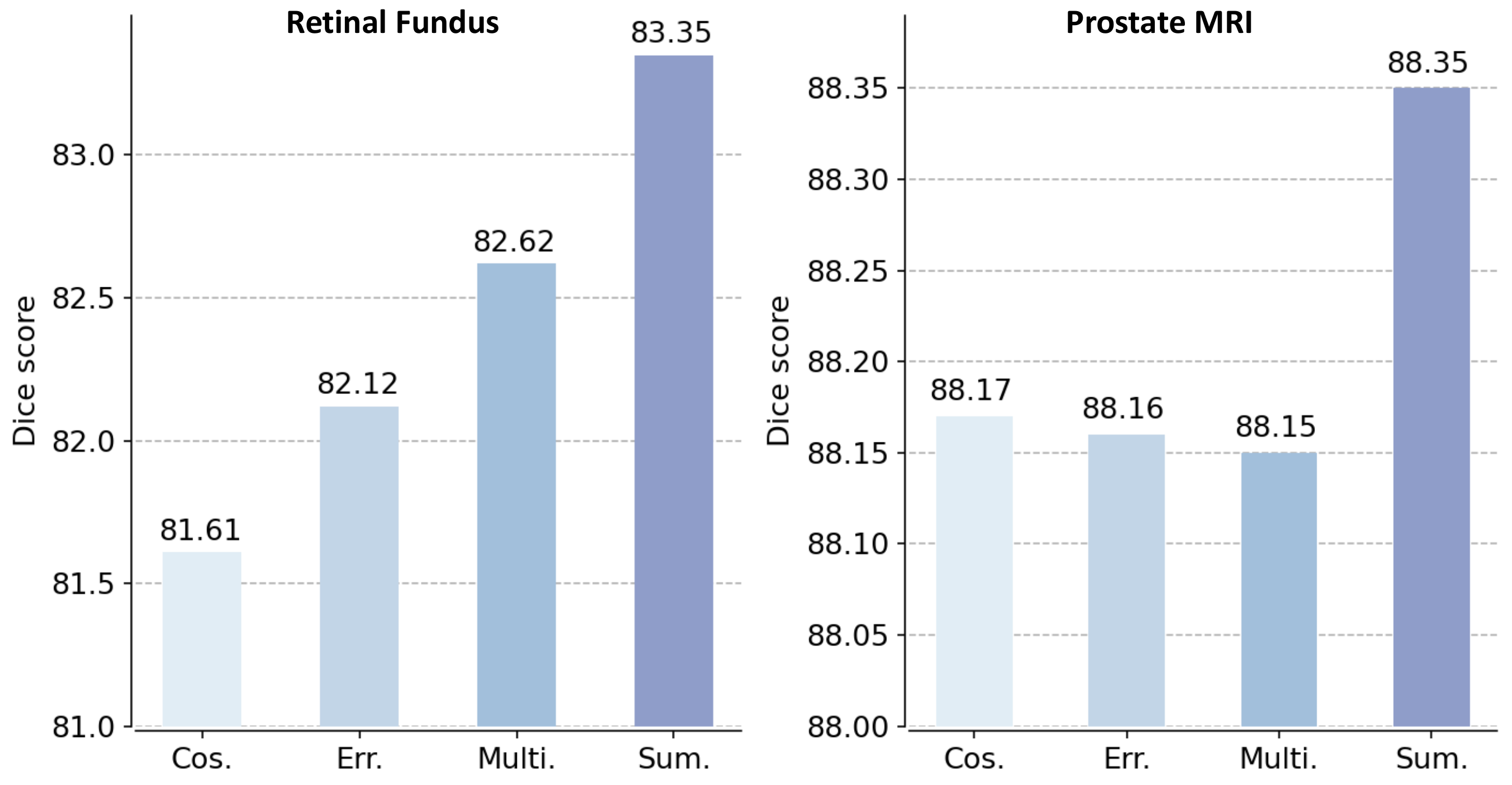} 
\vspace{-3mm}
\caption{Ablation study on effects of two separate contribution quantification metrics and their combination on two datasets.}
\label{fig:ablation_performance}
\vspace{-1.5mm}
\end{figure}
\\\textbf{Contribution of each component.}
We further conduct the ablation study for our two components (i.e., $\Gamma{(cos)}$ and $\Gamma{(err)}$).
As shown in Fig.~\ref{fig:ablation_performance}, solely using either one will lead to a decrease in the performance on both segmentation tasks. This is reasonable because either individual metric may not be able to fully quantify the contribution. As for the combination, the performance improvements reflect how the two components play complementary roles in improving the global model. Please note, on the prostate dataset, the performance differences by single measurement and multiplication are marginal. This may be because both follow similar trends for this application, where the differences between clients are significantly smaller than in the retinal dataset. Even in this case, the summation combination still shows improved performance.

\section{Conclusion}
We have studied a novel and practical problem of jointly tackling \textit{collaboration fairness} and \textit{performance fairness}. We have proposed a novel method to estimate client contributions from both gradient and data space, followed by a fair reward allocation based on those estimates. We further design a novel fair FL algorithm by using the estimated contributions to re-weight the global aggregation. Our solution provides inspiration to motivate more clients to join a FL project, leveraging larger and diverse data, benefiting the acceptance of FL for medical imaging and healthcare applications.  We conducted comprehensive experiments on two medical datasets and provided theoretical analysis for estimation robustness and model convergence. 
Our proposed estimation mechanism is extendable to other FL problems, such as Non-IID data and adversarial robustness. For future work, we plan to extend our method to detect adversarial clients and consider fairness on clients with noisy data.

\section*{Acknowledgement} 
\vspace{-1mm}
This work was supported by NVIDIA and National Natural Science Foundation of China (Project No. 62201485).

\clearpage
{\small
\bibliographystyle{unsrt}
\bibliography{egbib}
}

\clearpage
\appendix
\paragraph{Roadmap of Appendix:} 
The Appendix is organized as follows. We present theoretical proof of the robustness of distributional changes in Section~\ref{app:sec:proof_dis}, the proof of convergence in Section~\ref{app:sec:proof_conv}. Additional experiment results are in Section~\ref{app:sec:more_exp}.

\section{Proof of Value Difference Upper Bound}
\label{app:sec:proof_dis}
\subsection{Preliminary}
Given two distributions $\gD_s$ and $\gD_t$ over $\gZ$, let $\Pi_{st}$ denote the collection of joint distributions over $\gZ \times \gZ$. In particular, for all $\pi \in \Pi_{st}$, if iid draw $(s,t)\sim \pi$, then $s\sim\gD_s$ and $t\sim\gD_t$. Given a metric $d$ over $\gZ$, the Wasserstein distance is defined as the infimum over all such $\pi \in \Pi_{st}$ of the expected distance between $(s,t)\sim \pi$.
\begin{equation}
\label{app:eq:was_dis}
W_{1}\left(\mathcal{D}_{s}, \mathcal{D}_{t}\right) \triangleq \inf _{\pi \in \Pi_{s t}} \underset{(s, t) \sim \pi}{\mathbb{E}}[d(s, t)].
\end{equation}

\subsection{Assumptions and Proofs}
First, we state the assumption of Lipschitz stable, which is derived from a standard notation of deletion stability, often studied in the context of generalization~\cite{bousquet2002stability}. Following~\cite{distSV}, we assume our potential function is $B(k)$-Lipschitz stable.
\begin{assumption}
Let $(\gZ,d)$ be a metric space. For potential function $\Gamma$ and non-increasing $\gB: \sN \rightarrow [0,1]$, $\Gamma$ is $\gB$-Lipschitz stable with respect to $d$ if for all $k \in \sN$, $\gS \in \gZ^{k-1}$, and all $z,z' \in \gZ$,
\begin{equation}
\left|\Gamma(S, \{z\})-\Gamma\left(S,\left\{z^{\prime}\right\}\right)\right| \leq \gB \cdot d\left(z, z^{\prime}\right).
\end{equation}
\end{assumption}

For the convenience of notation, for any $z \in \gZ$ and subset $S \subseteq \gZ$, we denote $\Delta_z \Gamma(S) = \Gamma(S\backslash\{z\},\{z\})$. Therefore, fixing $z \in \gZ$, we can write $\hat{\nu}(z;\Gamma,\gD,N)$ as $\mathbb{E}_{\gS\sim\gD^N}[\Delta_{z}\Gamma(S)]$. 
Let $\pi \in \Pi_{st}$ be some coupling of $\gD_s$ and $\gD_t$, we reformulate this expectation as:
\begin{equation}
\begin{aligned} 
\underset{S \sim \mathcal{D}_{s}^{N}}{\mathbb{E}}\left[\Delta_{z} \Gamma(S)\right] 
=&\underset{S \times T \sim \pi^{N}}{\mathbb{E}}\left[\Delta_{z} \Gamma(S)\right] 
\\ 
=&\underset{S \times T \sim \pi^{N}}{\mathbb{E}}\left[\Delta_{z} \Gamma(S)-\Delta_{z} \Gamma(T)\right]
\\ &+\underset{S \times T \sim \pi^{N}}{\mathbb{E}}\left[\Delta_{z} \Gamma(T)\right]
\\ 
=&\underset{S \times T\sim \pi^{N}}{\mathbb{E}}\left[\Delta_{z} \Gamma(S)-\Delta_{z} \Gamma(T)\right]
\\&+\underset{T \sim \mathcal{D}_{t}^{N}}{\mathbb{E}}\left[\Delta_{z} \Gamma(T)\right],
\end{aligned}
\end{equation}
where the first and last equation follow our definition that the marginals of $\pi$ are $\gD_s$ and $\gD_t$, and the second equation follows by the linearity of expectation.

Then we bound the first term $\left[\Delta_{z} \Gamma(S)-\Delta_{z} \Gamma(T)\right]$. By expanding the difference between $\Delta_{z}\Gamma(S)$ and $\Delta_{z}\Gamma(T)$ into a telescoping sum of $N$ pairs of terms, we bound each pair to depend on a single draw $(s_i,t_i)\sim \pi$. For $S,T\in \gZ^N$, and $i \in \{0,\ldots,N\}$, denote $Z_{i}=\left(\bigcup_{j=i+1}^{N} s_{j}\right) \cup\left(\bigcup_{j=1}^{i} t_{j}\right)$, such that $\gZ_{0}=S$ and $\gZ_N = T$. Then we can expand the first term as:
\begin{equation}
\begin{aligned} 
\Delta_{z} \Gamma(S)\!-\!\Delta_{z} \Gamma(T)\!=\!\sum_{i=1}^{N} \Delta_{z} \Gamma\left(Z_{i-1}\right)-\Delta_{z} \Gamma\left(Z_{i}\right).
\end{aligned} 
\end{equation}
Since we assume $\Gamma$ is $\gB$-Lipschitz stable, we can derive the following bound:

\begin{equation}
\begin{aligned} 
\underset{S \times T \sim \pi^{N}}{\mathbb{E}}&\left[\Delta_{z} \Gamma(S)-\Delta_{z} \Gamma(T)\right] 
\\
&=\underset{S \times T \sim \pi^{N}}{\mathbb{E}}\left[\sum_{i=1}^{N} \Delta_{z} \Gamma\left(Z_{i-1}\right)-\Delta_{z} \Gamma\left(Z_{i}\right)\right] 
\\ &=\sum_{i=1}^{N} \underset{S, T \sim \pi^{N}}{\mathbb{E}}\left[\Delta_{z} \Gamma\left(Z_{i-1}\right)-\Delta_{z} \Gamma\left(Z_{i}\right)\right] 
\\ 
&=\sum_{i=1}^{N} \underset{\substack{(s_{i}, t_{i}) \sim \pi \\ R \in \mathcal{Z}^{N-2}}}{\mathbb{E}}\left[\Delta_{z} \Gamma\left(R \cup\left\{s_{i}\right\}\right)-\Delta_{z} \Gamma\left(R \cup\left\{t_{i}\right\}\right)\right] 
\\
& \leq 2 \gB \cdot \sum_{i=1}^{N} \underset{\left(s_{i}, t_{i}\right) \sim \pi}{\mathbb{E}}\left[d\left(s_{i}, t_{i}\right)\right] 
\\
& \leq 2N \gB \underset{(s, t) \sim \pi}{\mathbb{E}}[d(s, t)],
\end{aligned}
\end{equation}
where the last two inequality follow the $\gB$-Lipschitz assumption and the fact that each draw from $\pi$ is iid. 
Finally, we re-write the differences in values in terms of the infimum over $\Pi_{st}$ to complete the bound.
\begin{equation}
\begin{aligned}
\hat{\nu}&\left(z ; \Gamma, \mathcal{D}_{s}, N\right)-\hat{\nu}\left(z ; \Gamma, \mathcal{D}_{t}, N\right) 
\\
&\leq \inf _{\pi \in \Pi_{s t}} 
\left[\underset{S \times T \sim \pi^{N}}{\mathbb{E}}\left[\Delta_z \Gamma(S)-\Delta_{z} \Gamma(T)\right]\right] 
\\
& \leq 2N \gB \inf _{\pi \in \Pi_{s t}} \underset{(s, t) \sim \pi}{\mathbb{E}}[d(s, t)] 
\\ 
&= 2N \gB \cdot W_{1}\left(\mathcal{D}_{s}, \mathcal{D}_{t}\right)
\end{aligned}
\end{equation}

\section{Proof of FedCE Convergence}
\label{app:sec:proof_conv}
\subsection{Preliminary}
We start by setting up the basic FL training and objective. Then we give the proof of our theorem.

Let $\boldsymbol{G}_{(k, i)}$ denotes the locally accumulated stochastic gradients scaled with a factor $\gamma$. For the local client gradients and global model update, we have the following rule:
\begin{equation}
\label{app:eq:grad_update}
\left\{\begin{array}{l}\boldsymbol{G}_{(k, i)} \triangleq \frac{1}{\gamma_{(k, i)}} \sum_{\lambda=0}^{\kappa_{(k, i)}-1} \gamma_{(k,i)}^\lambda\nabla F_{i}\left(\boldsymbol{w}_{(k, i)}^{\lambda}\right) \\ \bs{w}_{k+1}-\bs{w}_{k}=-\eta  \boldsymbol{d}_{k},
\end{array}\right.
\end{equation}
where $
\boldsymbol{d}_{k} \triangleq \sum_{i=1}^{N} p_{i} \boldsymbol{G}_{(k, i)}
$, and $\kappa_{(k,i)}$ denotes the local update iterations (steps) for the client $i$ at the $k$-th round. $\gamma_{(k,i)}^\lambda$ denotes an arbitrary scalar, where $\bs \gamma_{(k,i)} = [\gamma_{k,i}^0, \cdots, \gamma_{k,i}^\lambda]$, $\gamma_{(k,i)} = \|\bs \gamma_{(k,i)}\|$, and we assume $\sum_{i=1}^N \frac{p_i}{\gamma_{(k, i)}\sqrt{\kappa_{(k,i)}}} \sum_{\lambda=0}^{\kappa_{(k, i)}-1} \gamma_{(k,i)}^\lambda = 1$ to make sure the summation of aggregation factors is 1 for each communication round.
For the global direction, we denotes it as the global gradient $\nabla \|F(\bs w_k)\|$. In particular, the global gradient in FL is the weighted average of all training clients, i.e.,
$\nabla \|F(\bs w_k)\| \triangleq \sum_{i=1}^N p_i \nabla F_i(\bs w_k)$, where $\nabla F_i(\bs w_k)$ is local gradient of $\bs w_k$ calculated on all training data from client $i$.
In FL, the learning objective is to find an optimal global model $\bs w_K^*$ by minimizing $F(\bs w_K^*)$, that is:
\begin{equation}
\boldsymbol{w}_{K}^{*} \triangleq \arg \min F\left(\boldsymbol{w}_{k}\right).
\end{equation}
In other words, the loss value of $F(\bs w_k)$ should decrease as training goes ($k$ increases). For the $k$-th round, we have the objective of:
\begin{equation}
\label{app:eq:local_obj}
\boldsymbol{w}_{k+1}^{*} \triangleq \arg \min \left\{F\left(\boldsymbol{w}_{k+1}^{*}\right)-F\left(\boldsymbol{w}_{k}\right)\right\}.
\end{equation}
By comparing Eq.(\ref{app:eq:grad_update}) and Eq.(\ref{app:eq:local_obj}), we have $\|\bs d_k\| \leq \|\nabla F(\bs w_k)\|$.

\subsection{Assumptions}
We first state the assumptions on local function smoothness and bounded gradients, which are commonly adopted in optimization literature~\cite{li2019convergence,fedprox,fedadam,scaffold,tong2020effective}.
\begin{assumption}
\label{app:assumption:1}
Each  local objective function is Lipschitz smooth, that is, for $k \in [0, K-1]$:
$$
\left\|\nabla F\left(\bs{w}_{k+1}\right)-\nabla F\left(\bs{w}_{k}\right)\right\| \leq L\left\|\bs{w}_{k+1}-\bs{w}_{k}\right\|
$$
\end{assumption}

\begin{assumption}
\label{app:assumption:2}
For any local gradient $\nabla F_i(\bs w_{(k,i)}^{\lambda})$ and $\lambda \in [0,\tau_{(k,i)}-1]$, there exists $\beta_{(k,i)}\geq0$, such that,
$$
\left\|\nabla F_{i}\left(\bs {w}_{k}\right)-\nabla F_{i}\left(\bs {w}_{(k, i)}^{\lambda}\right)\right\| \leq \beta_{(k, i)}\left\|\bs {w}_{k}-\bs {w}_{(k, i)}^{\lambda}\right\|
$$
\end{assumption}

\begin{assumption}
\label{app:assumption:3}
For all local gradients, $s \in [0,\lambda]$ and $\lambda \!\in\!  [1, \kappa_{(k,i)}\!-\!1]$, there exists constants $\delta_{(k,i)}\!\ge\! 0 $, such that,
$$
\left\|\sum_{s=0}^{\lambda-1} \nabla F_{i}\left(\bs {w}_{(k, i)}^{s}\right)\right\|^{2} \leq \delta_{(k, i)} \sum_{s=0}^{\lambda-1}\left\|\nabla F\left(\bs {w}_{k}\right)\right\|^{2}
$$
\end{assumption}

\subsection{Proof of the convergence theorem}
In this part, we show how to derive the convergence theorem. First, we start with the differences between $\bs w_{k+1}$ and $\bs w_k$. Since the global gradient is Lipschitz smooth, we have:

\begin{equation}
\label{app:eq:diff_init}
\begin{aligned}  F&\left(\boldsymbol{w}_{k+1}\right)-F\left(\boldsymbol{w}_{k}\right) 
\\ 
\leq &\nabla F\left(\boldsymbol{w}_{k}\right)\left(\boldsymbol{w}_{k+1}-\boldsymbol{w}_{k}\right)+\frac{L}{2}\left\|\boldsymbol{w}_{k+1}-\boldsymbol{w}_{k}\right\|^{2}
\\
= &-\eta \left\langle\nabla F\left(\boldsymbol{w}_{k}\right), \boldsymbol{d}_{k}\right\rangle+\frac{\eta^{2} L}{2}\left\|\boldsymbol{d}_{k}\right\|^{2}.
\end{aligned}
\end{equation}

The first inequality is from Lipschitz smooth assumption and the second equation is by inserting Eq.(\ref{app:eq:grad_update}). Then we reformulate the inner product term into the following form:
\begin{equation}
\label{app:eq:inner_change}
\begin{aligned} \left\langle\nabla F\left(\boldsymbol{w}_{k}\right), \boldsymbol{d}_{k}\right\rangle 
&= \frac{1}{2}\left\|\nabla F\left(\boldsymbol{w}_{k}\right)\right\|^{2}+\frac{1}{2}\left\|\boldsymbol{d}_{k}\right\|^{2}
\\&-\frac{1}{2} \|\nabla F(\bs w_k) - \bs d_k\|^2.
\end{aligned}
\end{equation}
By substituting Eq.(\ref{app:eq:inner_change}) into Eq.(\ref{app:eq:diff_init}), the inequation can be formulated as:

\begin{equation}
\begin{aligned}  
\label{app:eq:diff_obj}
F&\left(\boldsymbol{w}_{k+1}\right)-F\left(\boldsymbol{w}_{k}\right) 
\\ 
\leq &- \frac{1}{2} \eta
\left(\!\left\|\nabla F\left(\boldsymbol{w}_{k}\right)\right\|^{2}\!+\!\left\|\boldsymbol{d}_{k}\right\|^{2}
\!-\!\|\nabla F(\bs w_k) - \bs d_k\|^2 \!\right) 
\\
&+ \frac{\eta^{2} L}{2}\left\|\boldsymbol{d}_{k}\right\|^{2}
\\
=& -\frac{1}{2}\eta \|\nabla F(\bs w_k)\|^2 + \frac{\eta(\eta L-1)}{2} \|\bs d_k\|^2 
\\
& + \frac{\eta}{2} \|\nabla F(\bs w_k) - \bs d_k\|^2
\\
\leq & (\frac{\eta^2L}{2} - \eta)\|\nabla F(\bs w_k)\|^2 + \frac{\eta}{2} \|\nabla F(\bs w_k) - \bs d_k\|^2,
\end{aligned}
\end{equation}
when $\eta L - 1 \geq 0$. The last inequality is because $\|d_k\| \leq \|\nabla F(\bs w_k)\|$. Next, we present how to bound the term $\|\nabla F(\bs w_k) - \bs d_k\|^2$.

By the definition of $\bs d_k$, for $i \in [1,N]$ and $k \in [0,K-1]$, we have:
\begin{equation}
\label{app:eq:diff}
\begin{aligned}
\|\nabla F&(\bs w_k) - \bs d_k\|^2 = \left\|\nabla F\left(\boldsymbol{w}_{k}\right)-\sum_{i=1}^{N} p_{i} \boldsymbol{G}_{(k, i)}\right\|^{2}
\\
& =\left\|\sum_{i=1}^{N} p_{i}\left(\nabla F_{i}\left(\boldsymbol{w}_{k}\right)-\boldsymbol{G}_{(k, i)}\right)\right\|^{2}
\\
& \leq \sum_{i=1}^{N} p_{i}\left\|\nabla F_{i}\left(\boldsymbol{w}_{k}\right)-\boldsymbol{G}_{(k, i)}\right\|^{2}
\\
&=\sum_{i=1}^{N} p_{i}\left\|\nabla F_{i}\left(\boldsymbol{w}_{k}\right) \!-\!\frac{1}{\gamma_{(k, i)}} \!\!\!\!\sum_{\lambda=0}^{\kappa_{(k, i)}-1}\!\!\! \gamma_{(k,i)}^\lambda \nabla F_{i}\left(\boldsymbol{w}_{(k, i)}^{\lambda}\right)\right\|^{2} 
\\
&=\sum_{i=1}^{N} p_{i}\left\|\sum_{\lambda=0}^{\kappa_{(k, i)}-1}\!\! \frac{\gamma_{k,i}^\lambda}{\gamma_{(k, i)}}\left(\nabla F_{i}\left(\boldsymbol{w}_{k}\right)\!-\!\nabla F_{i}\left(\boldsymbol{w}_{(k, i)}^{\lambda}\right)\right)\right\|^{2} 
\\
&\leq \sum_{i=1}^{N} p_{i}\sum_{\lambda=0}^{\kappa_{(k, i)}-1} \frac{\gamma_{(k,i)}^\lambda}{\gamma_{(k, i)}}\left\|\nabla F_{i}\left(\boldsymbol{w}_{k}\right)-\nabla F_{i}\left(\boldsymbol{w}_{(k, i)}^{\lambda}\right)\right\|^{2}
\\
& \leq \sum_{i=1}^{N} p_{i} \sum_{\lambda=0}^{\kappa_{(k, i)}-1} 
\frac{\beta_{(k,i)}^2\gamma_{(k,i)}^\lambda}{\gamma_{(k, i)}} \left\|\bs w_k - \bs w_{(k,i)}^\lambda\right\|^2,
\end{aligned}
\end{equation}
where the first and second inequality uses Jensen's Inequality and the last inequality follows our assumption~\ref{app:assumption:2}. For training in FL, when local iteration $\lambda=0$, we have $\bs w_k = \bs w_{(k,i)}^\lambda$, this induces $
\|\bs w_k -\bs w_{(k,i)}^\lambda\|^2 = 0$ in Eq.(\ref{app:eq:diff}). So we consider the differences when $\lambda \geq 1$.

\begin{equation}
\label{app:eq:diff_w}
\begin{aligned} 
\left\|\boldsymbol{w}_{k}-\boldsymbol{w}_{(k, i)}^{\lambda}\right\|^{2}&=\eta^{2}\left\|\sum_{s=0}^{\lambda-1} \nabla F_{i}\left(\boldsymbol{w}_{(k, i)}^{s}\right)\right\|^{2} 
\\ 
\leq & \eta^{2} \delta_{(k, i)} \sum_{s=0}^{\lambda-1}\left\|\nabla F\left(\boldsymbol{w}_{k}\right)\right\|^{2}
\\
=& \eta^{2} \delta_{(k, i)} \lambda\left\|\nabla F\left(\boldsymbol{w}_{k}\right)\right\|^{2}.
\end{aligned}
\end{equation}
The inequality here follow our assumption~\ref{app:assumption:3}.
By inserting this equation back to Eq.(\ref{app:eq:diff}), we obtain:

\begin{equation}
\label{app:eq:diff_complete}
\begin{aligned}
\|\nabla F&(\bs w_k) - \bs d_k\|^2 
\\
&  \leq \sum_{i=1}^{N} p_{i} \sum_{\lambda=0}^{\kappa_{(k, i)}-1} 
\lambda\eta^2\frac{\beta_{(k,i)}^2 \delta_{(k,i)}\gamma_{(k,i)}^\lambda}{\|\bs\gamma_{(k, i)}\|} \left\| \nabla F(\bs w_k)\right\|^2
\\
& = \sum_{i=1}^{N} p_{i} \frac{\kappa_{k,i}(\kappa_{k,i}-1)}{2}\eta^2\beta_{(k,i)}^2 \delta_{(k,i)} \frac{\|\bs\gamma_{(k,i)}\|_1}{\|\bs\gamma_{(k,i)}\|}\|\nabla F(\bs w_k)\|^2.
\end{aligned}
\end{equation}
For the ease of notation, we define $\rho_{k,i} = \frac{\|\bs\gamma_{(k,i)}\|_1}{\|\bs\gamma_{(k,i)}\|\sqrt{\kappa_{(k,i)}}}$ and $A_{(k,i)}\!=\!\eta{\sqrt{\kappa_{(k,i)}
}(\kappa_{(k,i)}\!-\!1)}\beta_{(k,i)}^2 \delta_{(k,i)}$. Then we have:

\begin{equation}
\label{app:eq:diff_final}
\begin{aligned}
\|\nabla F&(\bs w_k) - \bs d_k\|^2 
\\
& \leq \frac{\eta}{2}\sum_{i=1}^{N} p_{i} \rho_{(k,i)} A_{(k,i)}\|\nabla F(\bs w_k)\|^2
\\
& = \frac{\eta}{2}\|\nabla F(\bs w_k)\|^2 \sum_{i=1}^{N} p_{i} \rho_{(k,i)} A_{(k,i)}.
\end{aligned}
\end{equation}
After obtaining the bound of the differences between server and normalized gradient, we are now ready to derive the final result.
Substituting Eq.(\ref{app:eq:diff_final}) into Eq.(\ref{app:eq:diff_obj}), we have:
\begin{equation}
\begin{aligned}  
\label{app:eq:conv_final}
F&\left(\boldsymbol{w}_{k+1}\right)-F\left(\boldsymbol{w}_{k}\right) 
\\ 
\leq & (\frac{\eta^2L}{2} - \eta)\|\nabla F(\bs w_k)\|^2 
\\
&+ \frac{\eta^2}{4}\|\nabla F(\bs w_k)\|^2 \sum_{i=1}^{N} p_{i} \rho_{(k,i)} A_{(k,i)}
\\
=& \left(\frac{\eta}{4}(2\eta L + \sum_{i=1}^{N} p_{i} \rho_{(k,i)} \eta A_{(k,i)}) - \eta\right)\|\nabla F(\bs w_k)\|^2.
\end{aligned}
\end{equation}

\subsection{Proof of the convergence corollary}
Here we further analyze relations between convergence and our reweighting factors to present the effects of our methods. Recall that in Eq.(\ref{app:eq:diff_obj}), $\eta L>1$. We also assume the summation of aggregation factors is 1.
Therefore, we can construct an inequation as below:
\begin{equation}
\label{app:eq:aux_ineq}
\begin{aligned}
\left(\eta\sum_{i=1}^N p_i A^{1/2}_{(k,i)} + \eta\sum_{i=1}^N p_i \rho_{(k,i)}L\right) \geq 1,
\end{aligned}
\end{equation}
where $\eta\sum_{i=1}^N p_i A^{1/2}_{(k,i)}$ is always positive.

Next, to ensure the model converge in Theorem~\ref{thm:convergence}, we need $\left(\frac{\eta}{4}(2\eta L + \sum_{i=1}^{N} p_{i} \rho_{(k,i)} \eta A_{(k,i)})-\eta\right)\leq 0$, that is,
$\left(\frac{1}{4}(2\eta L + \sum_{i=1}^{N} p_{i} \rho_{(k,i)} \eta A_{(k,i)})\right)\leq1$. By inserting Eq.(\ref{app:eq:aux_ineq}), we have:
\begin{equation}
\begin{aligned}
\frac{\eta}{4} & \left(2L+\sum_{i=1}^{N} \!p_{i} \rho_{(k,i)}A_{(k,i)}\right)
\\
& = \frac{\eta}{4} \sum_{i=1}^N \!p_i \left(2L\rho_{(k,i)} +\rho_{(k,i)}A_{(k,i)}\right)
\\
&\leq
\left(\eta\sum_{i=1}^N p_i A^{1/2}_{(k,i)} + \eta\sum_{i=1}^N p_i \rho_{(k,i)}L\right)
\\
& = \frac{\eta}{4}\sum_{i=1}^N p_i \left(4(A^{1/2}_{(k,i)}+\rho_{(k,i)}L) \right).
\end{aligned}
\end{equation}

To ensure this inequality always hold, we have:
\begin{equation}
\begin{aligned}
2L\rho_{(k,i)} + \rho_{(k,i)}A_{(k,i)} &\leq 
4(A^{1/2}_{(k,i)}+\rho_{(k,i)}L)
\\
\rho_{(k,i)}(A_{(k,i)}-2L) &\leq 4A^{1/2}_{(k,i)}
\\
\rho_{(k,i)} &\leq \frac{4A^{1/2}_{(k,i)}}{(A_{(k,i)}-2L)}
\\
&\text{(when $A_{(k,i)}-2L>0$)}
\end{aligned}
\end{equation}

We consider the convergence case when $A_{(k,i)}$ is dominant, then we have:
\vspace{-3mm}
\begin{equation}
\begin{aligned}
\rho_{(k,i)} \leq \frac{4A^{1/2}_{(k,i)}}{(A_{(k,i)}-2L)} = \mathcal{O}(\frac{1}{\sqrt{A_{(k,i)}}}).
\end{aligned}
\end{equation}

This indicates that the model converges when $\rho_{(k,i)}$ satisfy this condition. And we are able to minimize the upper bound of  $\rho_{(k,i)}$  by increasing $A_{(k,i)}$.

Recall that $A_{(k,i)} = \eta{\sqrt{\kappa_{(k,i)}
}(\kappa_{(k,i)}-1)}\beta_{(k,i)}^2 \delta_{(k,i)}$. 
The items $\eta$ and $\kappa_{(k,i)}$ are related with experimental settings. It is easy to understand that, if the training data are iid, increasing the learning rate $\eta$ or performing more local iterations $\kappa_{(k,i)}$ improves the convergence. For non-iid data, the convergence is also affected by data distribution. If we increase learning rate or local step, the model convergence speed may be improved at an early stage. However, it may let the model trap into a local optimum or suffer large client drifts when data are heterogeneous~\cite{scaffold}. 

Next we focus on terms of $\beta_{(k,i)}^2$ and $\delta_{(k,i)}$, which are related to our assumptions on the local gradients and parameters. 
According to Eq.(\ref{app:eq:diff_w}), we have:
\vspace{-3mm}
\begin{equation}
\begin{aligned}
\beta_{(k, i)}^2 &\geq \frac{\left\|\nabla F_{i}\left(\bs {w}_{k}\right)-\nabla F_{i}\left(\bs {w}_{(k, i)}^{\lambda}\right)\right\|^2}{\left\|\bs {w}_{k}-\bs {w}_{(k, i)}^{\lambda}\right\|^2}
\\
&\geq \frac{\left\|\nabla F_{i}\left(\bs {w}_{k}\right)-\nabla F_{i}\left(\bs {w}_{(k, i)}^{\lambda}\right)\right\|^2}
{\eta^{2} \delta_{(k, i)} \lambda\left\|\nabla F\left(\boldsymbol{w}_{k}\right)\right\|^{2}},
\end{aligned}
\end{equation}
which is also related to $\delta_{(k,i)}$. So we focus on discussing the relations between $\delta_{(k,i)}$ and convergence.
From the Assumption~\ref{app:assumption:3}, we have:
\vspace{-3mm}
\begin{equation}
\begin{aligned}
\delta_{(k, i)} \geq \frac{\left\|\sum_{s=0}^{\lambda-1} \nabla F_{i}\left(\bs {w}_{(k, i)}^{s}\right)\right\|^{2}}{\sum_{s=0}^{\lambda-1}\left\|\nabla F\left(\bs {w}_{k}\right)\right\|^{2}}.
\end{aligned}
\end{equation}
This term quantifies the percentage of local gradients over the global(aggregated) gradients. That is, to increase $\delta_{(k,i)}$, we need to weigh more on local gradients from client $i$. Since the client with boundary data or different distribution is under-represented during training, which harms the overall convergence. We need to assign higher weights to promote training on this kind of client, thus improving convergence. This well matches our contribution estimation method, i.e., allocating higher weight to clients presenting different information in gradient space or suffering high error on local data when their gradient is excluded.

\section{Additional Experimental Results}
\label{app:sec:more_exp}
\begin{figure}[t]
  \centering
   \includegraphics[width=0.99\columnwidth]{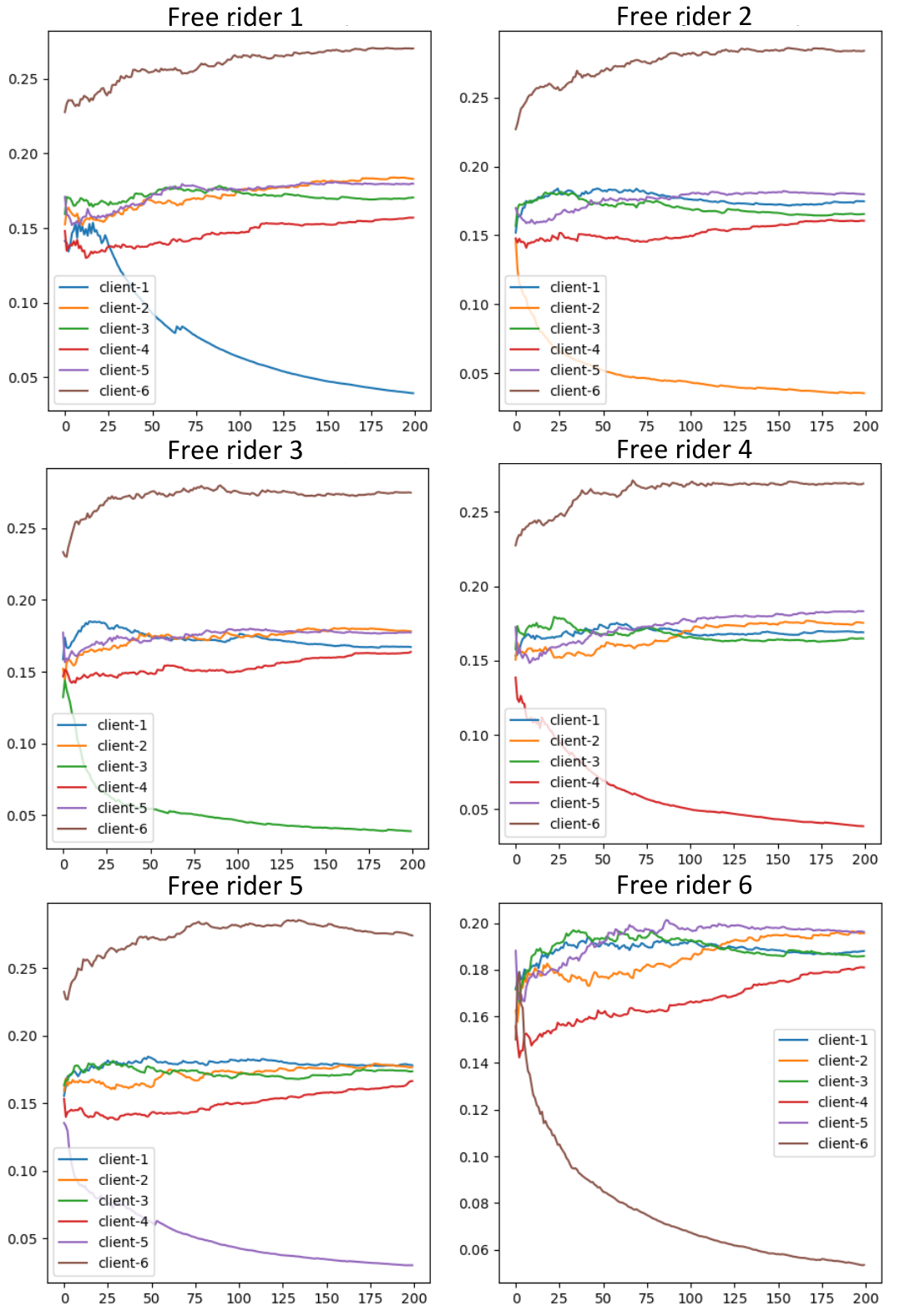}
   \vspace{-3mm}
   \caption{Free rider study by using cosine similarity between local and global gradients. X-axis denotes the communication rounds and y-axis denotes the similarity.}
   \label{app:fig:freerider_cos}
\end{figure}

In this section, we present more results of our method, including the free rider detection, discussion on client contributions, and visual comparison of segmentation results.
\paragraph{Free rider detection.}
We first present more results for the free rider detection using the prostate dataset. As discussed in the experiment section, we have combined the local-global gradients cosine similarity and local-global model error difference to detect the free rider client. Here we further present the results by using calculating the cosine similarity between local and global gradients, as shown in Fig.~\ref{app:fig:freerider_cos}. From the figure can be observed that the similarity between local gradients from the free rider and global clients decreases lower with training goes on. The free rider client can be distinguished within 50 rounds. Interestingly, we observe that client 6 presents a high cosine similarity, except itself is the free rider. This is because client 6 has more samples than other clients, and the gradients dominate others during the aggregation. Therefore, it is critical to combine both gradients and performance, which well matches our motivation for method design.
\begin{table*}[t!]
    \renewcommand\arraystretch{1.2}
    \centering
        \caption{\small{Client contribution quantification on the retinal fundus dataset by using performance drop with regard to leave-one-out experiments and using training sample proportions.}}
        \centering
        \vspace{-3mm}
        {
        \scalebox{0.88}{
        \setlength\tabcolsep{4pt}
        
        \begin{tabular}{@{}ccccccccccccccc@{}}
        \toprule
        Client                   & \multicolumn{2}{c}{1}       & \multicolumn{2}{c}{2}       & \multicolumn{2}{c}{3}       & \multicolumn{2}{c}{4}       & \multicolumn{2}{c}{5}       & \multicolumn{2}{c|}{6}              & \multicolumn{2}{c}{No} \\ \midrule
        Metric                   & Disc         & Cup          & Disc          & Cup         & Disc         & Cup          & Disc         & Cup          & Disc         & Cup          & Disc  & \multicolumn{1}{c|}{Cup}   & Disc       & Cup       \\ \hline
        Dice                     & 86.84        & 74.06        & 88.26         & 74.21       & 88.46        & 73.25        & 87.41        & 73.58        & 82.52        & 70.66        & 89.43 & \multicolumn{1}{c|}{74.05} & 89.43      & 75.50     \\
        $\Delta$ Dice                   & -2.59        & -1.44        & -1.17         & -1.29       & -0.97        & -2.25        & -2.02        & -1.92        & -6.91        & -4.84        & 0.00  & \multicolumn{1}{c|}{-1.45} & \multicolumn{2}{c}{-} \\ 
        Performance Contribution & \multicolumn{2}{c}{15.00\%} & \multicolumn{2}{c}{9.50\%}  & \multicolumn{2}{c}{12.00\%} & \multicolumn{2}{c}{15.00\%} & \multicolumn{2}{c}{44.00\%} & \multicolumn{2}{c|}{5.50\%}        & \multicolumn{2}{c}{-}   \\ \hline \hline
        Training Samples         & \multicolumn{2}{c}{50}      & \multicolumn{2}{c}{98}      & \multicolumn{2}{c}{47}      & \multicolumn{2}{c}{230}     & \multicolumn{2}{c}{80}      & \multicolumn{2}{c|}{400}           & \multicolumn{2}{c}{-}   \\
        Sample Contribution      & \multicolumn{2}{c}{5.52\%}  & \multicolumn{2}{c}{10.83\%} & \multicolumn{2}{c}{5.19\%}  & \multicolumn{2}{c}{25.41\%} & \multicolumn{2}{c}{8.84\%}  & \multicolumn{2}{c|}{44.20\%}       & \multicolumn{2}{c}{-}   \\ \bottomrule
        \end{tabular}

    }}
    \vspace{-3mm}
    \label{app:table:retinal_contribution}
\end{table*}
\begin{figure*}[t]
  \centering
   \includegraphics[width=0.99\textwidth]{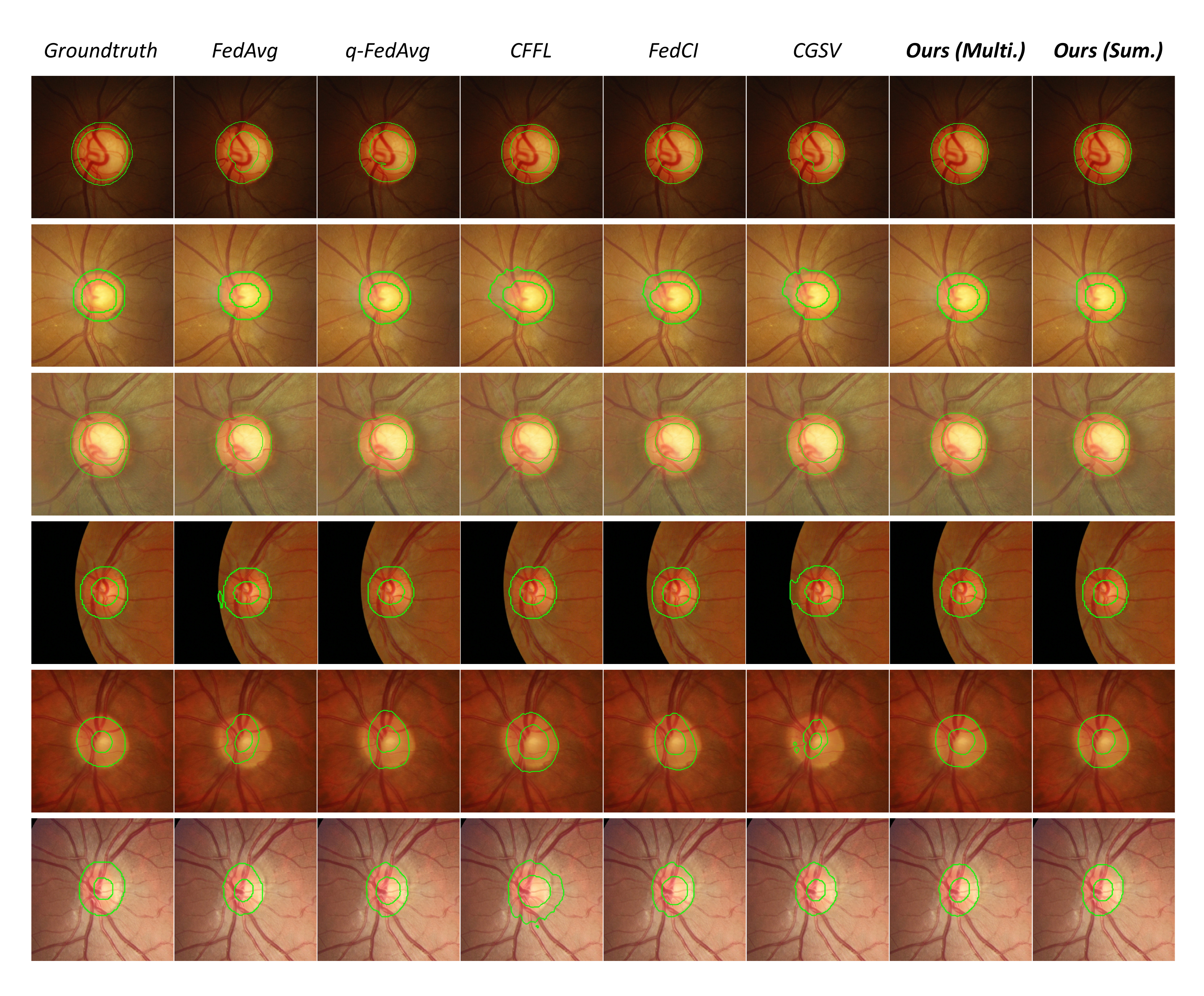}
   \vspace{-6mm}
   \caption{Qualitative comparison on the results of optic disc/cup segmentation from retinal fundus images. Each row denotes a client.}
   \label{app:fig:retinal_vis}
\end{figure*}
\begin{table*}[t!]
    \renewcommand\arraystretch{1.2}
    \centering
        \caption{\small{Client contribution quantification on the prostate dataset by using performance drop with regard to leave-one-out experiments and using training sample proportions.}}
        \centering
        \vspace{-3mm}
        {
        \scalebox{0.88}{
        \setlength\tabcolsep{5pt}
        \begin{tabular}{@{}ccccccc|c@{}}
        \toprule
        Client              & 1       & 2       & 3       & 4       & 5       & 6       & No       \\ \midrule
        Dice                & 84.90   & 87.95   & 87.91   & 87.97   & 76.67   & 87.53   & 88.32    \\
        $\Delta$ Dice          & -3.43   & -0.37   & -0.42   & -0.36   & -11.65  & -0.80   & -        \\
        Performance Contribution & 20.13\% & 2.19\%  & 1.74\%  & 2.09\%  & 68.47\% & 4.68\%  & -        \\\hline \hline
        Training Samples    & 381     & 238     & 278     & 242     & 389     & 814     & -    \\
        Sample Contribution           & 16.27\% & 10.16\% & 11.87\% & 10.33\% & 16.61\% & 34.76\% & -\\ \bottomrule
        \end{tabular}

    }}
    \vspace{-3mm}
    \label{app:table:prostate_contribution}
\end{table*}
\begin{table*}[ht]
    \renewcommand\arraystretch{1.0}
    \centering
        \caption{\small{Performance comparison using Dice score on image segmentation datasets of retinal fundus images and prostate MRI.}}
        \centering
        \vspace{-3mm}
        {
        \scalebox{0.88}{
        \setlength\tabcolsep{3 pt}
        \begin{tabular}{c|cccccccc|cccccccc}
    \toprule
             {Task}&\multicolumn{8}{c|}{Retinal Fundus Segmentation}&\multicolumn{8}{c}{Prostate MRI Segmentation}
              \\\cline{0-16}
            Client & 1 & 2 & 3 & 4 & 5 & 6 & Avg. & Std.
            &1 & 2 & 3 & 4 & 5 & 6 &Avg. & Std. \\
            \hline
            \multirow{2}{*}{Standalone} &
            86.69 & 85.51 & 86.21 & 89.91 & 
            79.77 & 90.98& 
            \multirow{2}{*}{86.51} & \multirow{2}{*}{3.95} & 
            91.23 & 84.59 & 87.57 & 87.37 & 
            86.70 & 89.25 & 
            \multirow{2}{*}{87.79} & \multirow{2}{*}{2.26}\\
            & $\pm$0.32 & {$\pm$1.41} & {$\pm$0.69} & {$\pm$0.15} & {$\pm$1.59} & {$\pm$0.06} & & &{$\pm$0.40} & {$\pm$0.55} & {$\pm$0.86} & {$\pm$0.32} & {$\pm$0.05} & {$\pm$0.13}\\ 
            \hline
            \multirow{2}{*}{FedAvg} &
            81.34 & 85.21 & 83.28 & 88.16 & 
            40.81 & 90.79 & 
            \multirow{2}{*}{78.27} & \multirow{2}{*}{18.66} & 
            
            91.10 & 84.59 & 89.02 & 89.09 & 
            83.87 & \textbf{89.27} & 
            \multirow{2}{*}{87.82} & \multirow{2}{*}{2.90}\\
            & {$\pm$3.08} & {$\pm$0.15} & {$\pm$1.60} & {$\pm$0.45} & {$\pm$6.57} & {$\pm$0.46} &&&
            {$\pm$0.10} & {$\pm$0.44} & {$\pm$0.37} & {$\pm$0.75} & {$\pm$0.42} & {$\pm$0.10}
            \\\hline
            \multirow{2}{*}{q-FedAvg} &
            86.24 & 86.97 & 87.37& 89.13 & 
            44.68 & 90.72 & 
            \multirow{2}{*}{80.85} & \multirow{2}{*}{17.80} & 
            
            90.94 & 85.60 & 89.28 & 89.18 & 
            84.27 & 88.67 & 
            \multirow{2}{*}{87.99} & \multirow{2}{*}{2.52}\\
            & {$\pm$0.80} & {$\pm$0.20} & {$\pm$0.66} & {$\pm$0.40} & {$\pm$3.42} & {$\pm$0.15} &&&
            {$\pm$0.25} & {$\pm$0.56} & {$\pm$0.37} & {$\pm$0.85} & {$\pm$0.32} & {$\pm$0.09}
            \\ \hline
            \multirow{2}{*}{CFFL} &
            85.72 & 86.29 & 86.96 & 88.62 & 
            41.12 & 90.16 & 
            \multirow{2}{*}{79.81} & \multirow{2}{*}{19.02} & 
            
            91.01 & 85.49 & 89.24 & 88.98 & 
            82.11 & 88.17 & 
            \multirow{2}{*}{87.50} & \multirow{2}{*}{3.20}\\
            & {$\pm$2.17} & {$\pm$1.32} & {$\pm$0.58} & {$\pm$1.95} & {$\pm$2.35} & {$\pm$0.95} &&&
            {$\pm$0.67} & {$\pm$0.72} & {$\pm$0.39} & {$\pm$0.86} & {$\pm$2.20} & {$\pm$0.41}
            \\ \hline
            \multirow{2}{*}{FedCI} &
            87.02 & 86.93 & 87.35 & 88.53 & 
            40.99 & 90.22 & 
            \multirow{2}{*}{80.17} & \multirow{2}{*}{19.24} & 
            
            91.21 & 85.40 & 89.49 & 88.37 & 
            83.96 & 88.49 & 
            \multirow{2}{*}{87.82} & \multirow{2}{*}{2.68}\\
            & {$\pm$1.47} & {$\pm$0.41} & {$\pm$0.40} & {$\pm$0.39} & {$\pm$7.94} & {$\pm$0.14} &&&
            {$\pm$0.68} & {$\pm$0.74} & {$\pm$0.57} & {$\pm$0.94} & {$\pm$0.49} & {$\pm$0.28}
            \\ \hline
            \multirow{2}{*}{CGSV} &
            83.46 & 85.57 & 85.47 & 88.48 & 
            33.79 & \textbf{91.01} & 
            \multirow{2}{*}{77.96} & \multirow{2}{*}{21.80} & 
            
            91.15 & 84.90 & 89.27 & 88.09 & 
            83.47 & 89.16 & 
            \multirow{2}{*}{87.67} & \multirow{2}{*}{2.91}\\
            & {$\pm$1.53} & {$\pm$0.15} & {$\pm$0.79} & {$\pm$0.71} & {$\pm$2.59} & {$\pm$0.65} &&&
            {$\pm$0.38} & {$\pm$0.66} & {$\pm$0.35} & {$\pm$0.93} & {$\pm$0.34} & {$\pm$0.25}
            \\
            \hline

            \multirow{2}{*}{FedCE (Multi.)} &
            86.73 & \textbf{87.45} & 87.51 & 89.26 & 
            \textbf{57.30} & 90.25 & 
            \multirow{2}{*}{\textbf{83.08}} & \multirow{2}{*}{\textbf{12.70}} & 
            
            \textbf{91.43} & \textbf{85.79} & 89.21 & 89.13 & 
            \textbf{85.68} & 88.62 & 
            \multirow{2}{*}{\textbf{88.31}} & \multirow{2}{*}{\textbf{2.22}}\\         
            & {$\pm$1.46} & {$\pm$0.14} & {$\pm$0.57} & {$\pm$0.32} & {$\pm$1.32} & {$\pm$0.15} &&&
            {$\pm$0.33} & {$\pm$0.55} & {$\pm$0.46} & {$\pm$0.59} & {$\pm$0.29} & {$\pm$0.10}
            \\ \hdashline
            \multirow{2}{*}{FedCE (Sum.)} & 
            \textbf{87.22} & 87.36 & \textbf{87.93} & \textbf{89.66} & 
            54.42 & 90.92 & 
            \multirow{2}{*}{82.92} & \multirow{2}{*}{14.03} & 
            
            91.18 & 85.54 & \textbf{89.59} & \textbf{89.22} & 
            84.99 & 88.79 & 
            \multirow{2}{*}{88.22} & \multirow{2}{*}{2.43} \\
            & {$\pm$0.61} & {$\pm$0.60} & {$\pm$0.56} & {$\pm$0.29} & {$\pm$1.84} & {$\pm$0.28} &&&
            {$\pm$0.30} & {$\pm$0.19} & {$\pm$0.33} & {$\pm$0.82} & {$\pm$0.44} & {$\pm$0.05}
            \\
       \bottomrule
        \end{tabular}
    }}
    \label{app:table:dice_res}
\end{table*}
\begin{figure*}[t]
  \centering
   \includegraphics[width=0.99\textwidth]{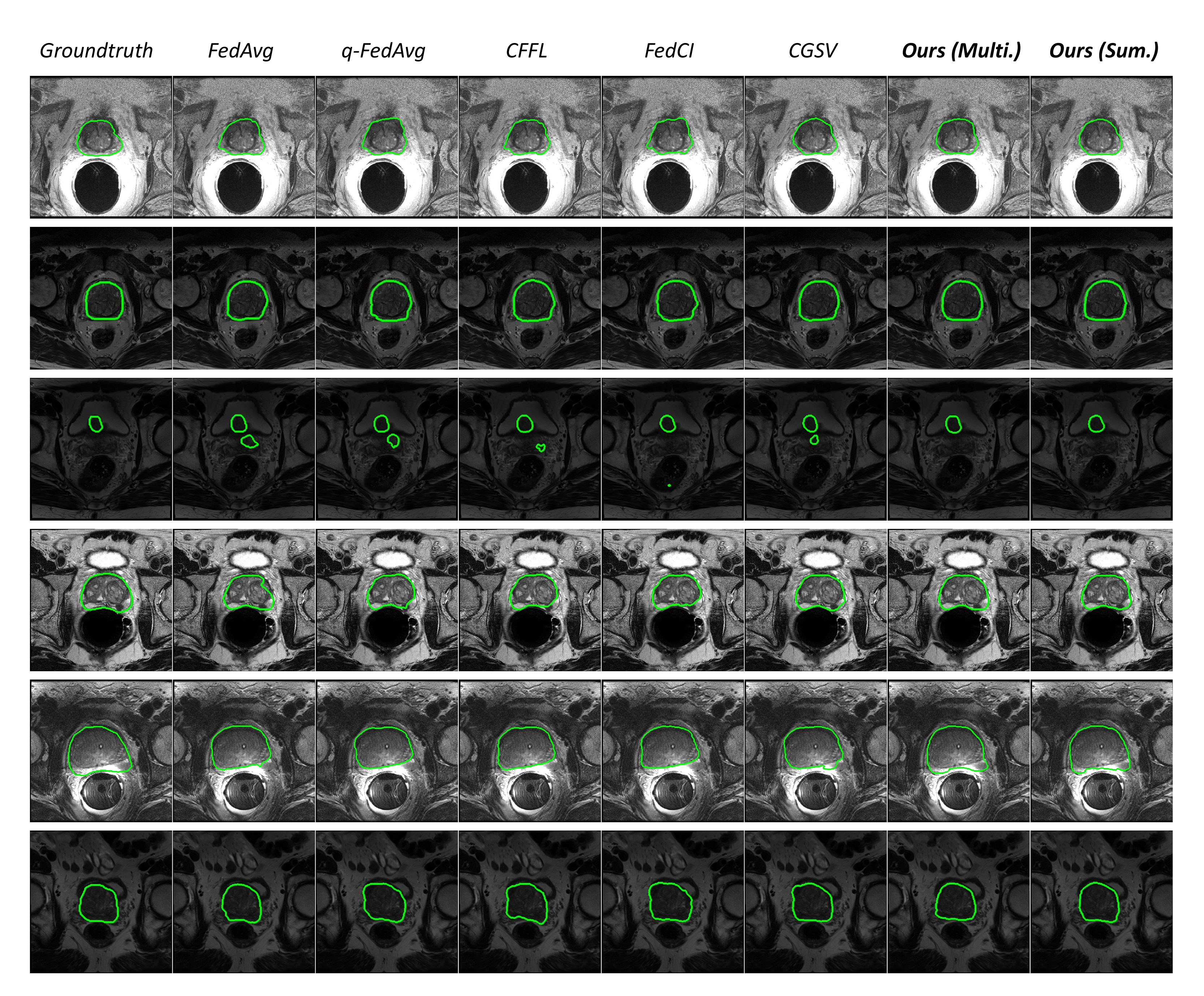}
   \vspace{-5mm}
   \caption{Qualitative comparison on the results of prostate segmentation. Each row denotes a client.}
   \label{app:fig:prostate_vis}
\end{figure*}
\paragraph{Client contribution quantification.}
We propose to quantify the client contribution by using the leave-one-out experiment, which is a popular and reliable way for data valuation~\cite{ghorbani2019data}. It assesses how much performance we will lose if we remove a certain client. However, it would be too computationally expensive to perform in practice. We hereby calculate the leave-one-out results as a reference to quantify client contribution in the context of performance. Specifically, we run six independent federated training by removing client $i \in \{1,\ldots,6\}$ to calculate the performance drop. Then we obtain the performance contribution by calculating the proportion of drop, i.e., a larger performance drop indicates this client has a larger performance contribution. Furthermore, in standard federated averaging algorithm~\cite{fedavg}, the sample proportion is typically used to indicate the importance (e.g., aggregation weight) of clients. So we calculate the sample contribution based on training samples. The results are shown in Table~\ref{app:table:retinal_contribution} and \ref{app:table:prostate_contribution}. From the two tables can be observed that, because the medical data collected from different sources are heterogeneous, the sample contribution does not strongly correlate with performance contribution, that is, more samples from one client may not improve the overall global performance a lot. This may be because some other clients with similar data distribution play a complementary role. For example, client 6 in the retinal dataset has over $40\%$ sample contribution, but the performance contribution is $5.5\%$ by the leave-one-out results. Therefore, solely considering the sample number is not enough if we aim to have a global model robust to various data distributions. In our experiments, we have presented how to promote collaboration fairness by considering the client contribution, which is reflected by client performance improvements. For the final client reward or credit allocation, it is a comprehensive procedure that needs to cover multiple different aspects, including our studies performance, as well as more factors like the computing cost, annotation cost, data quality, etc. The study on final client rewards or monetary allocation is still an open and important question that needs to be further investigated.
\paragraph{Distribution shifts on two datasets}
In this work, we consider two types of data heterogeneity sources to cover real medical scenarios. First is feature space shift from different imaging devices/protocols and variations during the imaging process, etc. In our scenario, prostate MRI data is captured by different machines and imaging protocols, and fundus image varies with different machines, illumination conditions, field of views, etc. The retinal dataset is ``less homogeneous'' than the prostate dataset because of more variations in color space and field of view.
Besides the feature shift, we also consider an additional special case shift, reflected by the retinal data: one of the clients has a different image setting (dual) from others (mono). This may not apply to most medical applications, hence is a ``less homogeneous'' data than most modalities.

\paragraph{Complete results with three random seeds}
We present the complete experiment results by reporting the mean and standard deviation of three independent runs in Table.~\ref{app:table:dice_res}. Notably, in the retinal fundus segmentation task, other compared methods exhibit a large standard deviation for the special client 5, while our method is more stable. Overall, our method yields stable results, demonstrating its reliability.

\paragraph{Visualization of segmentation results.}
We further present more qualitative segmentation results comparison on both retinal fundus dataset and prostate MRI dataset, as shown in Fig.~\ref{app:fig:retinal_vis} and Fig.~\ref{app:fig:prostate_vis}. In two figures, each row denotes one sample from a specific client, and each column denotes one method. We can see the samples visually looks different, showing the data heterogeneity of medical images collected from different hospitals/sources. Compared with alternative methods, which may present a less smooth boundary or cover more or less region, our methods (i.e., the multiplication and summation versions defined in Eq.~\ref{eq:comb_ways}.) present a more complete segmentation results with more accurate boundary and segmented region.

\end{document}

%% file: math_cmd.tex
\usepackage{amsmath,amsfonts,bm}

\def\eqref#1{equation~\ref{#1}}

\def\1{\bm{1}}

\DeclareMathAlphabet{\mathsfit}{\encodingdefault}{\sfdefault}{m}{sl}
\SetMathAlphabet{\mathsfit}{bold}{\encodingdefault}{\sfdefault}{bx}{n}

\def\gB{{\mathcal{B}}}

\def\gD{{\mathcal{D}}}
\def\gE{{\mathcal{E}}}

\def\gS{{\mathcal{S}}}

\def\gX{{\mathcal{X}}}
\def\gY{{\mathcal{Y}}}
\def\gZ{{\mathcal{Z}}}

\def\sN{{\mathbb{N}}}

\def\sR{{\mathbb{R}}}

\newtheorem{theorem}{Theorem}[section]

\newtheorem{definition}[theorem]{Definition}

\newtheorem{corollary}[theorem]{Corollary}

\newtheorem{assumption}[theorem]{Assumption}

\renewcommand{\hat}{\widehat}

\renewcommand{\bm}{\mathbbm}

\DeclareMathOperator{\Y}{{\mathcal Y}}

\newcommand{\bs}{\boldsymbol}